\documentclass{article} %
\usepackage[final]{neurips_2024}

\usepackage{amsmath,amsfonts,bm}

\def\eqref#1{equation~\ref{#1}}

\def\1{\bm{1}}

\DeclareMathAlphabet{\mathsfit}{\encodingdefault}{\sfdefault}{m}{sl}
\SetMathAlphabet{\mathsfit}{bold}{\encodingdefault}{\sfdefault}{bx}{n}

\usepackage{hyperref}
\usepackage{url}
\usepackage{microtype}
\usepackage{graphicx, floatrow}
\usepackage{subfig}
\usepackage{booktabs} %
\usepackage{etoolbox,xspace}
\usepackage{hyperref}
\usepackage{amsmath}
\usepackage{amssymb}
\usepackage{mathtools}
\usepackage{amsthm}
\usepackage{wrapfig}
\usepackage{caption}  %
\usepackage{enumitem}
\usepackage{titlesec}
\usepackage{cleveref}
\usepackage{multirow}
\usepackage{tcolorbox}
\usepackage{xcolor}
\usepackage{lineno}
\usepackage{longtable}
\usepackage{tikz}
\usepackage{soul}

\definecolor{darkblue}{rgb}{0, 0, 0.5}
\hypersetup{colorlinks=true, citecolor=darkblue, linkcolor=darkblue, urlcolor=darkblue}

\crefformat{section}{\S#2#1#3}
\crefname{figure}{Fig.}{Figs.}
\crefname{appendix}{App.}{Apps.}

 \titlespacing\section{1pt}{1pt plus 1pt minus 1pt}{1pt plus 1pt minus 1pt}
 \titlespacing*{\paragraph}{0pt}{0.5ex plus 0.2ex minus 0.1ex}{0.5em}

\definecolor{orchid}{RGB}{218,112,214}   %
\definecolor{purple}{RGB}{128,0,128}      %
\definecolor{tabblue}{RGB}{31,119,180}    %
\definecolor{green}{RGB}{0,128,0}         %
\definecolor{tabolive}{RGB}{128,128,0}    %
\definecolor{taborange}{RGB}{255,127,14}

\newcommand{\highlightred}[1]{{\tiny \textcolor{red}{(#1 \%)}}}

\newcommand{\model}{\textsc{ThinkPRM}\xspace}

\title{Process Reward Models That Think}

\author{
\begin{tabular}{c}
\textbf{Muhammad Khalifa}$^{\heartsuit}$, 
\textbf{Rishabh Agarwal}$^{\spadesuit}$, 
\textbf{Lajanugen Logeswaran}$^{\diamondsuit}$,\\
\textbf{Jaekyeom Kim}$^{\diamondsuit}$, 
\textbf{Hao Peng}$^{\clubsuit}$, 
\textbf{Moontae Lee}$^{\diamondsuit}$, 
\textbf{Honglak Lee}$^{\heartsuit\diamondsuit\star}$, 
\textbf{Lu Wang}$^{\heartsuit\star}$
\end{tabular} \\[0.5em]
\begin{tabular}{c}
$^{\heartsuit}$University of Michigan \quad
$^{\spadesuit}$Mila\\
$^{\diamondsuit}$LG AI Research \quad
$^{\clubsuit}$University of Illinois Urbana-Champaign\\[0.25em]
$^{\star}$Equal supervision
\end{tabular} \\
\texttt{\href{mailto:khalifam@umich.edu}{khalifam@umich.edu}}
}

\begin{document}

\maketitle

\begin{abstract}

Step-by-step verifiers—also known as process reward models (PRMs)---are a key ingredient for test-time scaling, but training them requires expensive step-level supervision. This work aims to build data-efficient PRMs as verbalized step-wise reward models that verify every step in the solution by \emph{generating} a verification chain-of-thought (CoT). We propose {\model}, a long CoT verifier fine-tuned on orders of magnitude fewer process labels than those required by discriminative PRMs. 
Our approach capitalizes on the inherent reasoning abilities of long CoT models, and outperforms LLM-as-a-Judge and discriminative verifiers---using only 1\% of the process labels in PRM800K---across several challenging benchmarks. Specifically, \model beats the baselines on ProcessBench, MATH-500, and AIME ’24 under best-of-N selection and reward-guided search. In an out-of-domain evaluation over subsets of GPQA-Diamond and LiveCodeBench, our PRM surpasses discriminative verifiers trained with the full PRM800K by 8\% and 4.5\%, respectively. Lastly, under the same token budget, \model scales up verification compute more effectively compared to LLM-as-a-Judge, outperforming it by 7.2\% on a subset of ProcessBench. This work highlights the value of generative, long CoT PRMs that can scale test-time compute for verification while requiring minimal supervision for training.\footnote{Our code, data, and models are released at \url{https://github.com/mukhal/thinkprm}.}

\end{abstract}
\vspace{-1.em}
\begin{figure}[ht!]
    \centering
    \includegraphics[width=0.85\linewidth]{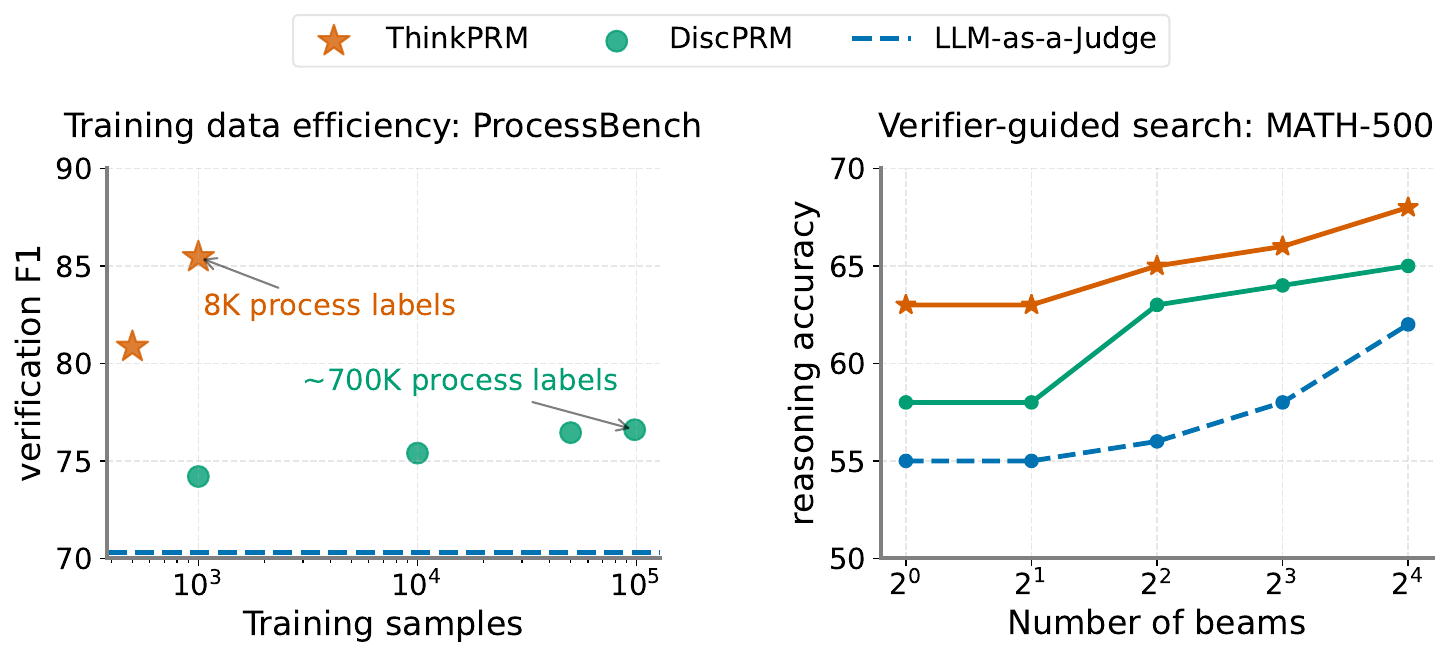}
    \caption{
    \textbf{Left:} Verifier F1-score on ProcessBench~\citep{zheng2024processbench}. \model-14B, trained on 8K process labels or 1K synthetic examples, outperforms discriminative PRMs trained on about 100x more data. 
    \textbf{Right:} Verifier-guided search accuracy on MATH-500 with Llama-3.2-3B-Instruct as generator. \model-1.5B, trained using the same 8K labels, outperforms LLM-as-a-judge and discriminative verifiers in reward-guided search on MATH-500. The LLM-as-a-judge in both figures uses the same base model as \model.
    }
    \label{fig:teaser}
\end{figure}

\section{Introduction}

Reasoning with large language models (LLMs) can substantially benefit from utilizing more test-time compute \citep{jaech2024openai,guo2025deepseek,akyurek2024surprising}. This typically depends on a high-quality process reward model (PRM)---also known as a process verifier---that scores (partial) solutions for selecting promising paths for search or ranking \citep{gsm8k,li2023making,wu2024inference,brown2024large}. %
PRMs have typically assumed the form of discriminative classifiers, trained to discern correct from incorrect reasoning \citep{uesato2022solving,prmlessons}. However, training discriminative PRMs requires access to process labels, i.e., step-level annotations, which either require extensive human annotation \citep{lightman2023let,zheng2024processbench}, gold step-by-step solutions \citep{grace2023}, or compute-intensive rollouts \citep{luo2024improve,chen2024alphamath}. For instance, training reasonably performing math PRMs requires hundreds of thousands of step-level annotations \citep{lightman2023let,math-shepherd}.

Generative verification either via LLM-as-a-judge~\citep{gpteval1,liu2023g,zheng2023judging} or GenRM \citep{zhang2024generative} treats verification as a generation problem of a rationale followed by a decision. 
However, LLM-as-a-judge is known to perform poorly compared to specialized reward models~\citep{lambert2024rewardbench,zhang2024small,chen2024tree}, as general-purpose LLMs frequently fail to recognize reasoning errors~\citep{huang2023large,zhang2024generative,ye2024physics}. 
Moreover, GenRM is limited to outcome verification via \textit{short} chain-of-thoughts (CoTs), fundamentally limiting its ability for test-time scaling.

\begin{wrapfigure}{r}{0.40\textwidth}
  \centering
  \includegraphics[width=0.99\linewidth]{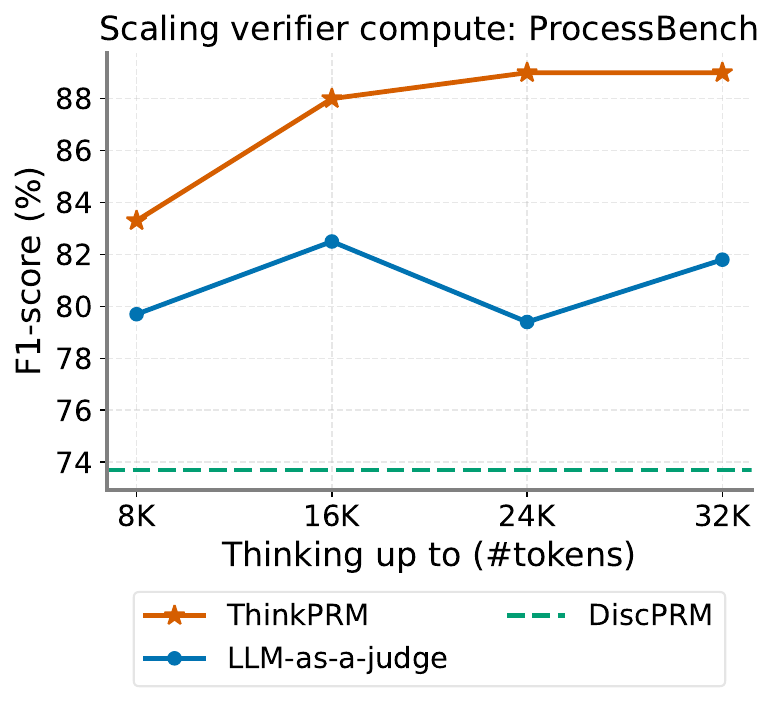}
  \caption{\model enables scaling verification compute with more CoT tokens.}
  \label{fig:verifier-compute}
\end{wrapfigure}

This paper builds on the insight that generative step-by-step verification can greatly benefit from scaling up the verifier’s inference compute—specifically, by enabling it to \textit{think} through a CoT. Specifically, we repurpose open-weight large reasoning models (LRMs) as the foundation for generative PRMs through \textit{lightweight} training. This training uses uses synthetic data \citep{kim2023prometheus,zhu2023judgelm,wang2024self}, utilizing as few as 8K step labels, and yieldinga \model—a PRM that not only surpasses LLM-as-a-judge, but also outperforms discriminative PRMs trained on two orders of magnitude more data across a variety of test-time scaling scenarios.

We obtain \model by training four reasoning models, namely R1-Distill-Qwen\{1.5B,7B,14B\} \citep{guo2025deepseek}, and QwQ-32B-Preview \citep{qwq-32b-preview}, and extensively evaluate it both as a standalone verifier on ProcessBench \citep{zheng2024processbench}, and combined with a generator under Best-of-N and verifier-guided beam search. \model-14B outperforms a discriminative PRM based on the same base model in terms of accuracy while using far fewer supervision signals as in \Cref{fig:teaser} left. In addition, \model-1.5B demonstrates strong performance on MATH-500 \citep{hendrycks2021measuring} under guided beam search, shown in \Cref{fig:teaser} right. Lastly, as shown in  \Cref{fig:verifier-compute}, \model can effectively utilize more verification compute than LLM-as-a-judge, by forcing it to think for more tokens. All these results are obtained while training only on 8K step labels.

Our work highlights the promise of long CoT PRMs that \textit{verify reasoning with reasoning}, effectively scaling both generator and verifier compute. Our main findings are as follows: \model outperforms strong PRM baselines in best-of-N and guided-search setups on two math reasoning benchmarks: MATH-500 and AIME 2024, and surpasses LLM-as-a-judge baselines under the same base model by thinking longer during verification (\cref{sec:main-results}). Moreover, \model generalizes under two types of domain shift. First, it outperforms baselines on out-of-domain tasks such as scientific reasoning and code generation. Second, despite being trained only on short solutions, it generalizes to long-form reasoning without explicit step delimiters (\cref{sec:qwen3-ver}). Third, \model outperforms self-consistency~\citep{wang2022self} when using the same compute budget, especially under high sampling regimes (\cref{sec:compute-matched}). Finally, fine-grained filtering of synthetic data based on step supervision is crucial for training high-quality PRMs (\cref{sec:filtering-ablation}).

\section{Background and Related Work}
\paragraph{Discriminative PRMs.}
Discriminative PRMs are trained as classifiers that directly predict numerical correctness scores for each solution step, and typically rely on extensive step-level annotations~\citep{uesato2022solving, lightman2023let, prmlessons}. Given a solution prefix, discriminative PRMs encode the solution text and employ a classification head to produce step-level scores, usually optimized with binary cross-entropy. An overall correctness score for a solution is obtained by aggregating these step-level scores~\citep{beeching2024scalingtesttimecompute}. PRMs are effective and straightforward but they do not utilize the language-modeling head of the base language model, making training expensive and labor-intensive~\citep{yuan2024free}. Additionally, they offer limited interpretability and utilize \textit{fixed} compute, restricting their dynamic scalability at test-time~\citep{zhang2024generative, mahan2024generative}. Thus, there is a need for data-efficient PRMs that can scale with more test-time compute.

\paragraph{Generative Verification.}
Generative verification ~\citep{zheng2023judging, zhu2023judgelm,zhang2024generative} frames verification as a language-generation task, producing step-level decisions as tokens (e.g., “correct” or “incorrect”), typically accompanied by a chain-of-thought (CoT). One can train generative verifiers using the standard language modeling objective on verification rationales rather than on binary labels. %
This approach leverages the strengths of LLMs in text generation, making generative verifiers inherently interpretable and scalable~\citep{zhang2024generative, mahan2024generative, gpteval1,ankner2024critique}. However, prior work on generative verifiers has relied mainly on short verification CoT (e.g., few hundred tokens) \citep{zhang2024generative}, which highly limits their scalability. Thus, there is a need for verifiers that can ``think'' longer through verification, utilizing test-time compute effectively. 
While LLM-as-a-Judge has been employed for step-level verification~\citep{zheng2024processbench}. it tends to be sensitive to prompt phrasing, and prone to invalid outputs, such as infinite looping or excessive overthinking~\citep{bavaresco2024llms}---issues we further confirm in this work. Prior results with reasoning models like QwQ-32B-Preview~\citep{qwq-32b-preview} show promise, but their practical utility in test-time scaling remains limited without additional training~\citep{zheng2024processbench}.

\paragraph{Test-Time Scaling with PRMs.}
Test-time scaling techniques, such as Best-of-N selection~\citep{charniak2005coarse} and tree-based search~\citep{yao2023tree, chen2024tree, wan2024alphazero}, leverage additional inference-time compute to improve reasoning performance. Central to these approaches is the quality of the verifier used to score and select solutions.  A major advantage of generative PRMs is that they uniquely support simultaneous scaling of both generator and verifier compute ~\citep{zhang2024generative, kalra2025verdict}. In particular, our work shows that generative PRMs trained based on long CoT models \citep{jaech2024openai,guo2025deepseek} enable both parallel and sequential scaling of verifier compute.

\section{\model}
\label{sec:model}

\begin{figure}
    \centering
    \includegraphics[width=0.90\linewidth]{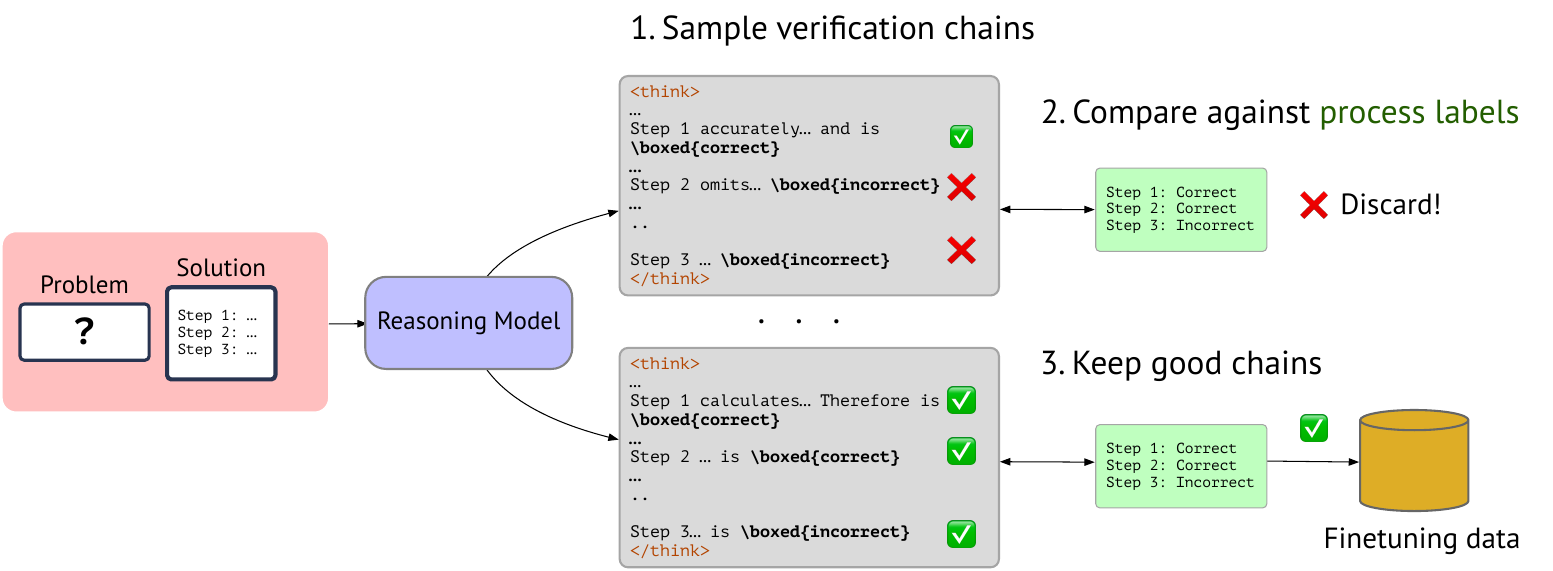}
    \caption{Collecting verification chains for finetuning. First, we prompt a reasoning model, in our case QwQ-32B-Preview to critique a given solution to a problem. Then, we sample multiple verification chains, which we judge against gold process labels from PRM800K, only keeping chains that match the gold process labels.}
    \label{fig:data-collection}
    
\end{figure}

Our goal is verbalized PRM that, given a problem-solution pair, verifies every step in the solution via an extended chain-of-thought (CoT) such as the one shown in \Cref{fig:thinkprm-example-verification} in \Cref{app:thinkprm-examples}.  This section introduces issues with LLM-as-a-judge verification and proposes a data collection process (shown in \Cref{fig:data-collection}) to curate high-quality synthetic verification CoTs for training such PRM. The rest of the paper addresses the following research questions: 

\begin{itemize}[leftmargin=*,noitemsep,topsep=0pt,parsep=0pt,partopsep=0pt]
\item \textbf{RQ1:} How well do LRMs perform under LLM-as-a-judge for process-level verification? \Cref{sec:rq1}
\item \textbf{RQ2:} Can lightweight finetuning on synthetic verification CoTs improve the reliability and effectiveness of these models as process verifiers? \Cref{sec:rq2}
\item \textbf{RQ3:} How does a finetuned verbalized PRM (\model) compare to discriminative PRMs and LLM-as-a-Judge baselines under different test-time scaling scenarios? \Cref{sec:main-results}
\end{itemize}

\subsection{LLM-as-a-judge PRMs are suboptimal}
\label{sec:rq1}

This section highlights limitations we observe when using off-the-shelf reasoning models as process verifiers, suggesting the need for finetuning. %
For evaluation, we use ProcessBench \citep{zheng2024processbench}, which includes problem-solution pairs with problems sourced from existing math benchmarks, along with ground-truth correctness labels.  
We report the binary F1-score by instructing models to verify full solutions and judge whether there exists a mistake. We use two most challenging subsets of ProcessBench: OlympiadBench \citep{he2024olympiadbench} and OmniMath \citep{gao2024omni}, each comprised of 1K problem-prefix pairs. For LLM-as-a-judge, we use the same prompt template as in \citet{zheng2024processbench}, shown in \Cref{tab:zeroshot-prompt}, which we found to work best overall. \Cref{tab:pb-results-detailed} shows LLM-as-a-judge F1 scores and a sample output by QwQ-32B-Preview is displayed in \Cref{tab:qwq-zeroshot-example-out} in \Cref{app:laj-details}.

We observe different issues with LLM-as-a-judge verification. First, the verification quality is \textit{highly sensitive} to the instruction wording: slight change in the instruction can affect the F1-score by up to 3-4 points.
First, a substantial number of the generated chains include \textit{invalid judgments}, i.e., chains without an extractable overall label as clear in \Cref{fig:invalid-labels}. Such invalid judgements are caused by the following. In some cases, final decision was in the wrong format than instructed e.g., the model tries to \textit{solve} the problem rather than verify the given solution---a behavior likely stemming from the model training. 
Second, we noted multiple instances of \textit{overthinking} \citep{chen2024not,cuadron2025danger}, which prevents the model from terminating within the token budget, and \textit{infinite looping/repetitions}, where the model gets stuck trying alternative techniques to verify the solutions. %

\Cref{fig:pb-length-analysis} (left) shows a histogram of verification CoT lengths generated by R1-Qwen-14B in the LLM-as-a-judge setting. Accurate CoTs tend to be shorter, typically under 3K tokens, while inaccurate CoTs are more evenly distributed and spike sharply around 7K-8K tokens, highlighting the prevalence of overthinking and looping in long chains.
We show examples of these behaviors in \Cref{app:issues}. 
In the next section, we mostly fix these issues via lightweight finetuning over synthetic verification CoTs.

\subsection{Finetuning on synthetic data boosts LLM-as-a-judge verification}
\label{sec:rq2}

\label{sec:finetuning}

Inspired by recent work on reducing overthinking in long CoT models that by training \citep{yu2024distilling,kang2024c3ot}, we aim to improve LLM-as-a-judge performance via finetuning on high-quality verification data. Collecting real data would be expensive, so we rely on filtered synthetic data \citep{zelikman2022star,singh2023beyond,dong2023raft,zhang2024small,wang2024self} also known as rejection sampling finetuning. To keep our approach simple, we refrain from more expensive training techniques, such as reinforcement learning or preference-based learning.

\paragraph{Synthetic data collection.}

As training data, we sample synthetic verification CoTs from QwQ-32B-Preview, prompting it to verify each step in a solution prefix, using the instruction shown in \Cref{fig:prompt-data-generation}. The problems and corresponding step-by-step solutions come from the PRM800K dataset \citep{lightman2023let}, which provides both model-generated solutions and human-verified step-level labels.

\begin{figure}[t]
    \centering
    \includegraphics[width=0.90\linewidth]{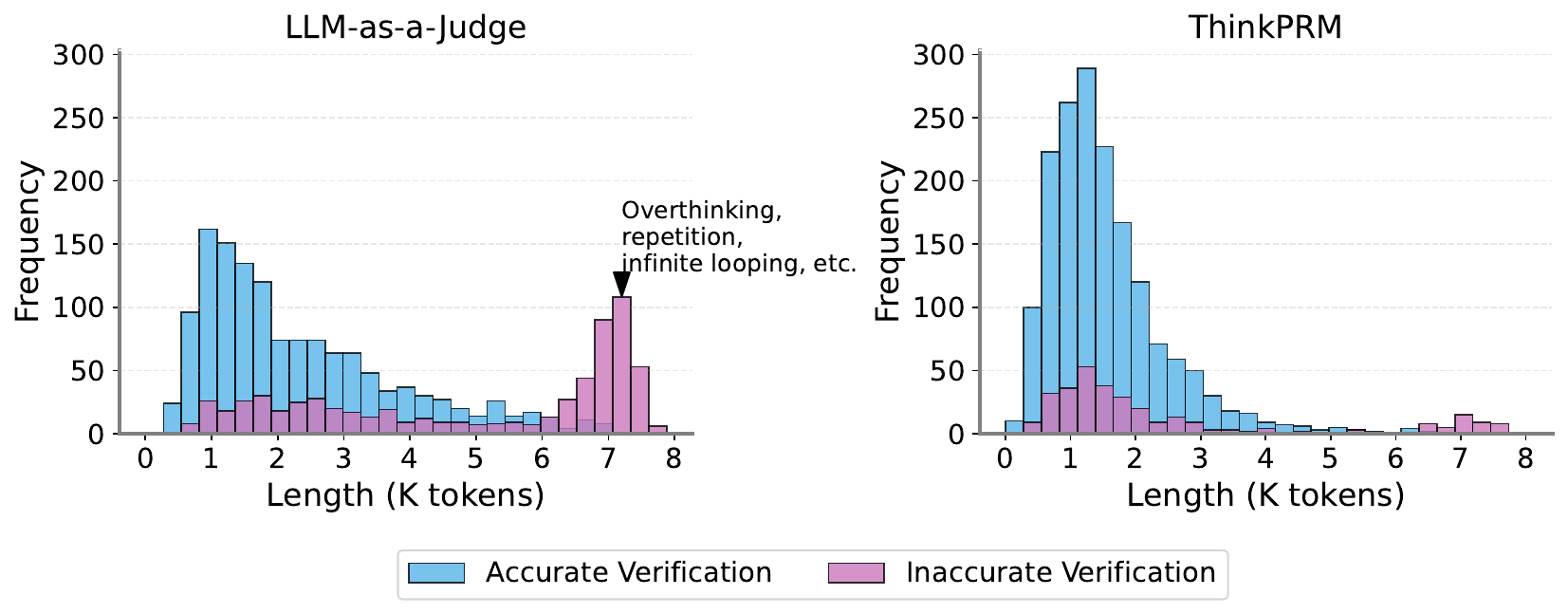}
    \caption{
    Verifier performance on ProcessBench in light of CoT lengths. On the left, LLM-as-a-judge produces excessively long chains including repetition, infinite looping, and overthinking, leading to worse verifier performance since the output never terminates. Training on collected syntehtic data substantially reduces these issues as shown in the \model plot on the right.}
    
    \label{fig:pb-length-analysis}
\end{figure}

The sampling process continues until we obtain 1K verification CoTs which coreepond to 8K step labels in total. For data filtering, we use the following criteria: \textbf{(i)} the CoT must follow the expected format (i.e., include an extractable decision label for each step inside \texttt{\textbackslash boxed\{\}} as shown in \Cref{fig:train-example}, and \textbf{(ii)} the generated step judgements match the gold step labels from PRM800K, and \textbf{(iii)} the CoT length is within a maximum budget---to avoid the excessive overthinking behavior we observed in \Cref{fig:pb-length-analysis} (left). The filtering process ensures our training data is of sufficient quality. note that process-based filtering is crucial for the performance of the resulting PRM as we show in \Cref{sec:filtering-ablation}.
Data collection is illustrated in \Cref{fig:data-collection}, data statistics are in \Cref{app:data-collection} and a training example is in \Cref{fig:train-example}.

Notably, our filtering relies only on step-level annotations, not on gold verification rationales or CoTs—making this pipeline scalable and low-overhead. %
In the absence of gold step-level annotations, one can obtain silver labels via Monte Carlo rollouts \citep{math-shepherd,chen2024alphamath}. 
While we train only on math data, the resulting PRM remains robust under other domains such as science QA and code generation as we show in \Cref{sec:results}. We then proceed to train our models on the 1K collected chains. Our training is very lightweight; finetuning QwQ-32B-Preview takes only 4.5 hours on a single A100 80GB GPU. Refer to \Cref{app:model-training} for training details.

\begin{figure}[h!]
  \centering
  \begin{floatrow}
    \ffigbox[\FBwidth]
      {\includegraphics[width=0.90\linewidth]{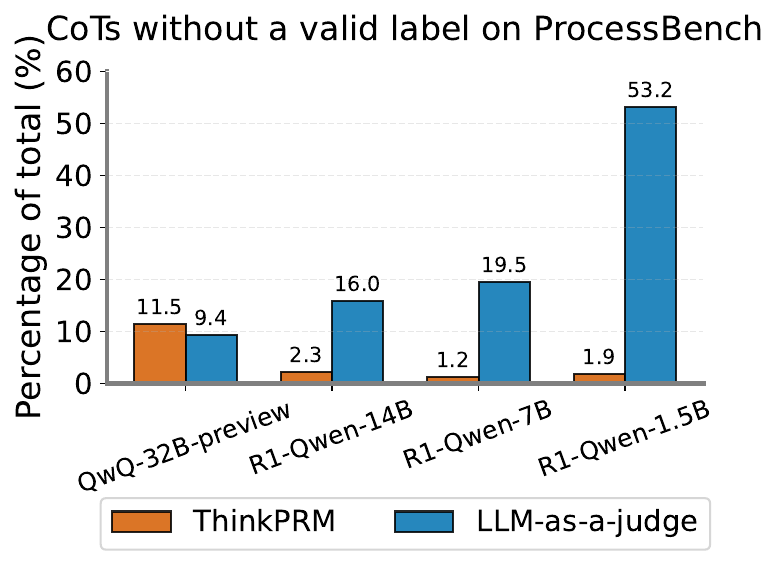}
      }
      {

      \caption{LLM-as-a-judge suffers from a significant ratio of verification CoTs that do not terminate with a parsable label, i.e., \texttt{\textbackslash boxed\{yes\}} or \texttt{\textbackslash boxed\{no\}}. Our finetuning process that yields \model, substantially mitigates this issue. Both verifiers are based on R1-Distill-Qwen-14B. } 
                          \label{fig:invalid-labels}
      }
    \ffigbox[\FBwidth]
      {
      \includegraphics[width=0.90\linewidth]{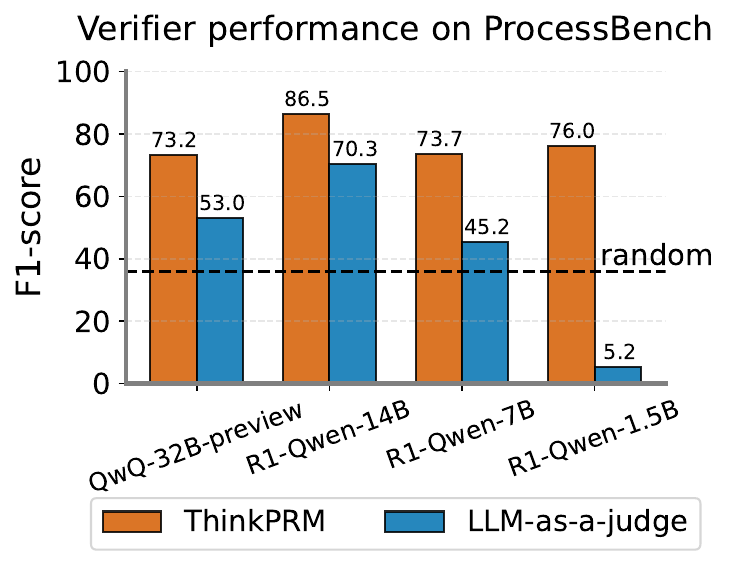}}
      {\caption{Verification accuracy on 2K question-solution pairs from two most challenging subsets of ProcessBench: OlympiadBench and OmniMath. \model obtained by finetuning the correponding model over only 1K verification chains performs better.}
        \label{fig:f1-pb}

      }
  \end{floatrow}
  \vspace{-0.4cm}
\end{figure}

\paragraph{Finetuning on synthetic verification CoTs substantially improves the verifier.}
\model trains on the 1K chains and is evaluated on ProcessBench and compared to LLM-as-a-judge under the same base model. \Cref{fig:f1-pb} shows verifier accuracy of different models before and after our finetuning. We note a substantial boost in F1 across all models, with the 1.5B model gaining most improvement by over 70 F1 points, and the 14B model performing best. Looking at the ratio of invalid judgements in \Cref{fig:invalid-labels}, we also note a significant reduction in invalid labels with all models, except for QwQ, where it slightly increases. Lastly, the reduction in overthinking and infinite looping behavior discussed in the last section is evident, as in \Cref{fig:pb-length-analysis} (right), where \model generations maintain a reasonable length (1K-5K) tokens while being substantially more accurate.

\section{Test-time Scaling Experiments}
\label{sec:main-results}

This section aims to answer RQ3 introduced in \cref{sec:model} by comparing \model to baselines under different scaling scenarios. We study how \model performs under different generation budgets  \textbf{(i)} best-of-N selection \citep{wu2024inference,brown2020language} and \textbf{(ii)} guided beam search \citep{snell2024scaling,beeching2024scalingtesttimecompute}. We also explore how \model performs when verifier compute is scaled either in parallel by aggregating decisions over multiple verification CoTs or sequentially through longer CoTs by forcing the model to double check or self-correct its verification.

\subsection{Experimental Setup}
\label{sec:exp-setup}
In the remainder of the the paper, we will mainly use our finetuned verifiers based on R1-Distill-Qwen-1.5B and R1-Distill-Qwen-14B as these provide the best tradeoff between size and performance. We will refer to these as \model-1.5B and \model-14B, respectively. 

\paragraph{Baselines.} We compare \model to \textbf{DiscPRM}, which uses the same base model as \model, finetuned with binary cross-entropy on the \textit{entire} PRM800K dataset, totaling 712K process labels, which is two orders of magnitude larger than our training data. Details on finetuning DiscPRMs are in \Cref{app:disc-training}. We also compare to \textbf{unweighted majority voting}, which merely selects the most frequent answer across the samples \citep{wang2022self}, and to LLM-as-a-Judge using the same base model as \model, prompted as in \Cref{sec:rq1}.

\paragraph{Tasks and Models.} We show results on three math reasoning tasks, namely 100 problems from MATH-500 \citep{hendrycks2021measuring} covering all difficulty levels (see \Cref{app:math-100} for more details), and American
Invitational Mathematics Examination (AIME) problems for 2024. Since \model was finetuned only on math data, we study the out-of-domain generalization on two tasks: scientific reasoning and code generation. For scientific reasoning, we use the physics subset of GPQA-Diamond \citep{rein2024gpqa}, consisting of 86 PhD-level multiple choice questions. For code generation, we use a 200-problem subset from the v5 release of LiveCodeBench \citep{jain2024livecodebench}.

Over MATH-500, we show results with \model-1.5B and \model-14B on two different generator models: Qwen-2.5-14B and Llama-3.2-3B-Instruct. The former model is used for best-of-N and the latter for beam search as search is compute intensive. Showing results with different generators guarantees that our conclusions are not specific to a certain model family or size. For the more challenging tasks, namely AIME '24 and GPQA, we use a more capable model, namely Qwen-2.5-32B-Instruct. For code generation, we use Qwen-2.5-Coder-7B \citep{qwencoder}. Implementation and hyperparemter details on how we select the final answer with best-of-N and beam search are in \Cref{app:eval-details}.

 \begin{figure}[t!]
    \centering
    \includegraphics[width=0.90\linewidth]{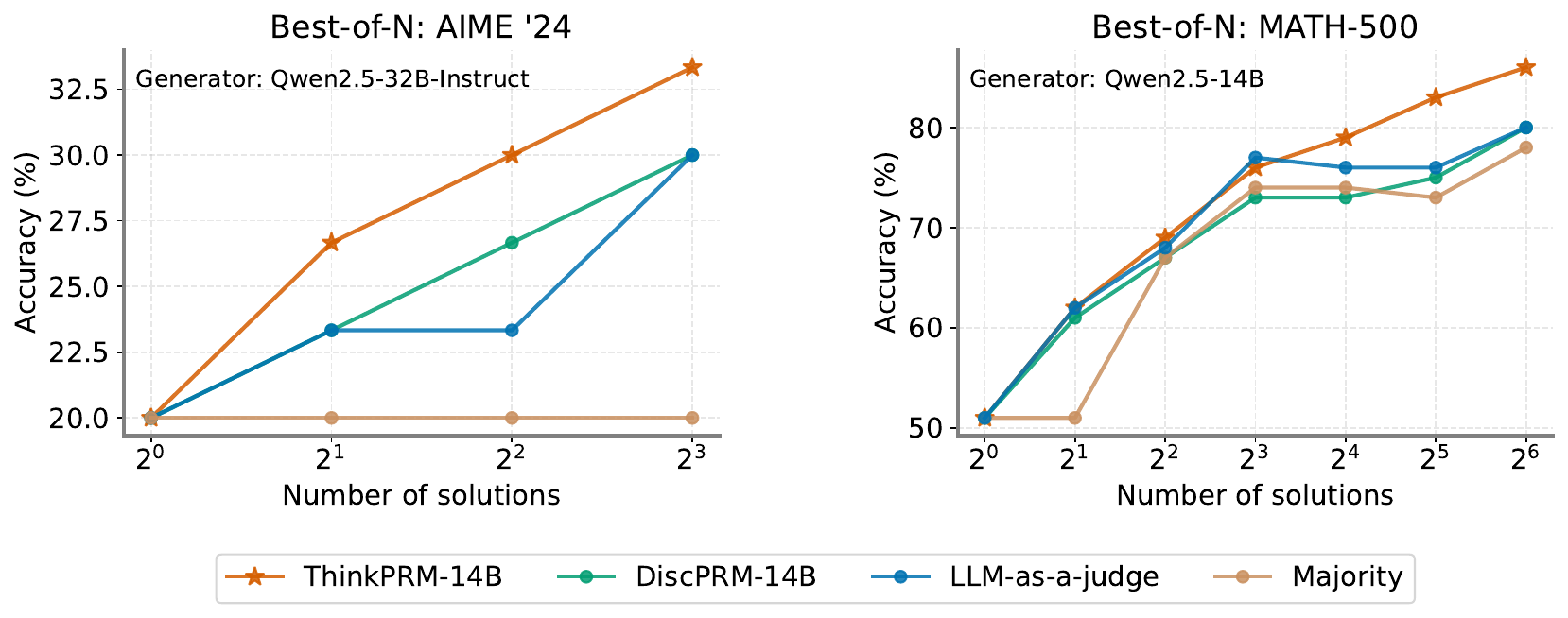}
    \caption{Best-of-N on AIME '24 and MATH-500. Compared to LLM-as-a-judge, DiscPRM, and (unweighted) majority vote, \model-14B exhibits best accuracy scaling curve.}
    \label{fig:aime-math-best-of-n}
\end{figure}

\paragraph{Scaling verifier compute.} Compared to DiscPRMs, generative reward models enable an extra dimension of scaling to squeeze more performance: scaling the verifier compute. Specifically,\model allows for two types of scaling. First, we use \textit{parallel scaling} \citep{mahan2024generative,brown2024large}, by sampling $K$ independent CoTs and averaging their scores. We will refer to this scaling using ``@K'' throughout the rest of the paper. 
Second, and more specific to long reasoning models, we use \textit{sequential scaling} e.g., by enabling the model to double-check its initial verification \citep{xiong2025self,kumar2024training,ye2024physics}. Inspired by \cite{muennighoff2025s1}, we use a trigger phrase such as \textit{``Let's verify again''} to elicit self-correction of earlier verification. See \Cref{sec:bf} for more details. %

\begin{figure}[t!]
  \centering
  \begin{floatrow}
    \ffigbox[\FBwidth]
      {\includegraphics[width=0.95\linewidth]{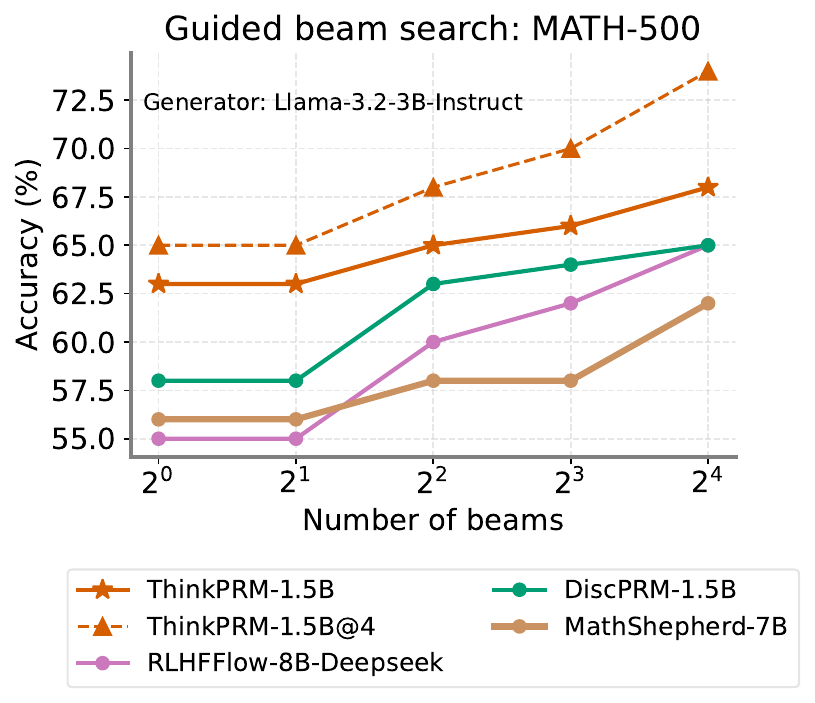}}
      {\caption{
      Comparison to off-the-shelf PRMs trained on much more step labels than \model. $@K$ represents parallel scaling by averaging scores over K CoTs.}\label{fig:beam-search-math500}}
    \ffigbox[\FBwidth]
      {\includegraphics[width=0.95\linewidth]{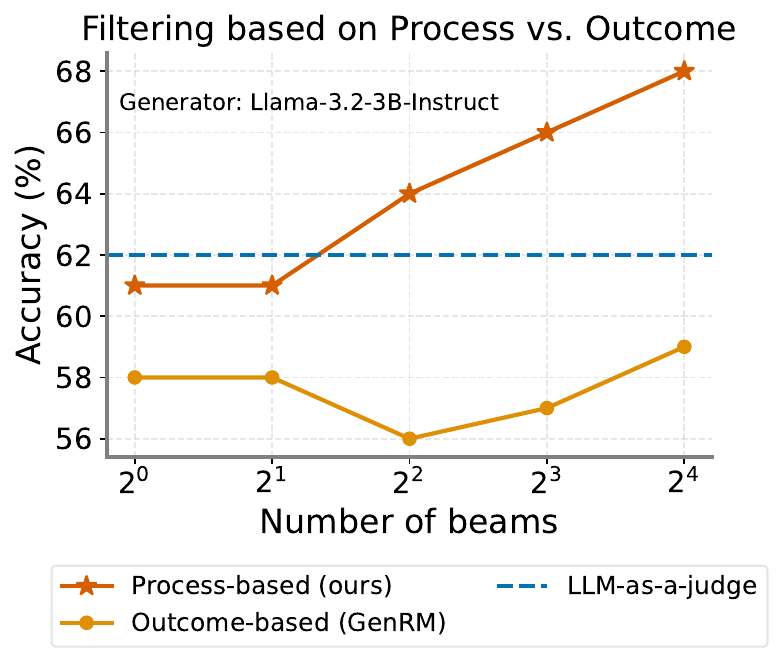}}
      {\caption{Ablating the data filtering mechanism, where our process-based filtering yields better PRMs. LLM-as-a-judge is shown with number of beams = 16.}\label{fig:process-vs-outcome}}
  \end{floatrow}
\end{figure}

\subsection{Results}
\label{sec:results}
\paragraph{\model outperforms DiscPRM and LLM-as-a-Judge.} Under best-of-N selection with MATH-500 shown in \Cref{fig:aime-math-best-of-n} (right), \model leads to higher or comparable reasoning accuracy to DiscPRM under all sampling budgets. The trend holds on the more challenging AIME '24, shown in \Cref{fig:aime-math-best-of-n} left. Additionally,
\Cref{fig:teaser} (right) shows beam search results on MATH-500, with \model 1.5B surpassing DiscPRM and LLM-as-a-Judge.

\paragraph{\model surpasses off-the-shelf PRMs.}
We compare \model-1.5B to two strong off-the-shelf PRMs, namely RLHFFlow-Deepseek-PRM \citep{xiong2024rlhflowmath} and MATH-Shepherd-PRM \citep{math-shepherd}. These PRMs are trained on even more data than PRM800K and are larger than 1.5B. We show results under verifier-guided search on MATH-500 in \Cref{fig:beam-search-math500}, with \model-1.5B's scaling curve surpassing all baselines and outperforming RLHFFlow-Deepseek-PRM, the best off-the-shelf PRM among the ones we tested, by more than 7\% across all beam sizes.

\begin{figure}[t!]
    \centering
    \includegraphics[width=0.85\linewidth]{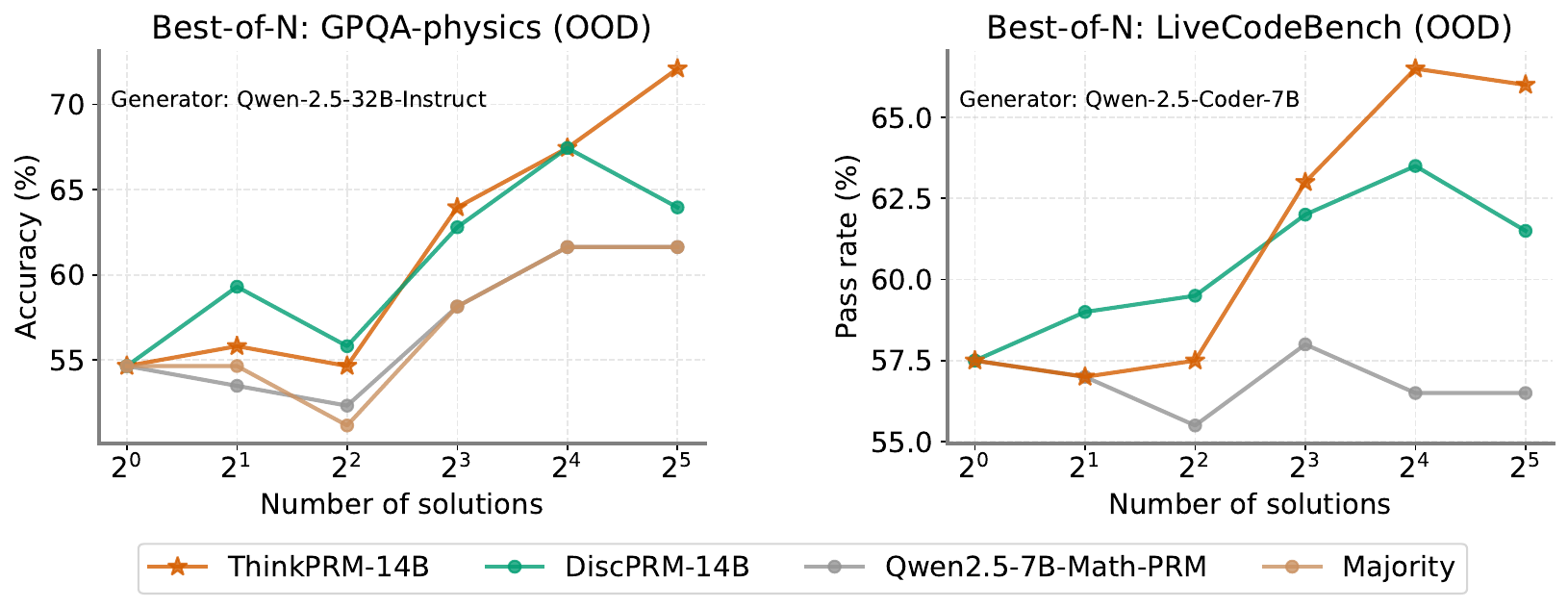}
    \caption{Best-of-N on two out-of-domain tasks: science QA (GPQA-Physics) and code generation (LiveCodeBench). Although \model was only finetuned on math, it exhibits superior OOD performance than the baselines, especially at larger sampling budgets, where the baselines fall short. Discriminative PRMs struggle despite being trained on orders of magnitude more process labels.}
    \label{fig:ood-best-of-n}
\end{figure}

\paragraph{\model excels on out-of-domain tasks.}
As for OOD performance on GPQA-physics (\Cref{fig:ood-best-of-n} left), \model scales better than DiscPRM---which drops substantially at N=32---outperforming it by 8\%. On LiveCodeBench  (\Cref{fig:ood-best-of-n} right), \model also outperforms DiscPRM by 4.5\%.  On LiveCodeBench, Qwen2.5-7B-Math-PRM \citep{prmlessons}---a discriminative PRM trained on substantial amount of process labels obtained from LLM-as-a-judge data and Monte Carlo rollouts---struggles when applied out-of-domain. Our results shed light on the fragility of discriminative PRMs under domain shifts in contrast with generative PRMs.

\paragraph{Scaling \model compute boosts performance.} Under verifier-guided search (shown in \Cref{fig:beam-search-math500}), parallel scaling with \model-1.5B@4 boosts the accuracy by more than 5\% points, and yields the best accuracy on MATH-500. In addition, parallel scaling with \model-14B@4 and \model-14B@8 boosts best-of-N performance on MATH-500 as shown in \Cref{fig:parallel-scaling-math} in \Cref{app:scaling-verifier-compute}. Now we move to sequential scaling of verifier compute by forcing \model to recheck its own verification. Since this can be compute-intensive, we only run this on 200 problems from OmniMath subset of ProcessBench, and observe how verification F1 improves as we force the model to think for longer as shown in \Cref{fig:verifier-compute}. \model exhibits better scaling behavior compared to LLM-as-a-judge, which drops after 16K tokens, and outperforms DiscPRM-14B by 15 F1 points. In summary, \model is consistently better than LLM-as-a-judge under parallel and sequential scaling.

\paragraph{Parallel scaling vs. sequential scaling.} Is it preferable to scale verifier compute in parallel or sequentially? We investigate this by comparing the two modes of scaling under the same token budget. \Cref{fig:parallel-vs-sequential} in \Cref{app:scaling-verifier-compute} shows performance of best-of-N with Qwen-2.5-14B under parallel and sequential scaling with $K=2,4$ under both parallel scaling and sequential scaling. Overall, the performance of both methods is fairly close, but we observe a slight advantage to parallel scaling under certain budgets.

\section{Analysis and Discussion}
\vspace{0.4em}
\label{sec:discussion}

\subsection{Training data efficiency}
\label{sec:data-efficiency}

A major strength of \model is training data efficiency compared to discriminative versions. Here, we study the training scaling behavior of \model-14B by training it over 500 and 1K examples in total collected using the pipeline in \Cref{sec:finetuning}, which roughly corresponds to 4K and 8K process labels from PRM800K in total. We compare that to DiscPRM-14B trained with 1K, 10K, 50K and 98K examples, where 98K corresponds to training on the full PRM800K train set that includes 712K step labels. \Cref{fig:teaser} (Left) contrasts the training data scaling behavior of \model-14B with that of DiscPRM-14B, where \model-14B's performance scales substantially better with two orders of magnitude fewer process labels. This primarily stems from \model's utilization of text generation and reasoning abilities of the underlying models.

While we train \model using only 1K data points, we investigate whether it will benefit from training on more data. Using the pipeline, we collect and filter additional verification CoTs and obtain a total of 65K chains. We then finetune R1-Distill-Qwen-1.5B and R1-Distill-Qwen-14B on these for a single epoch while keeping all other training hyperparameters fixed. We then compare the resulting models to the 1K-trained version of \model under best-of-N selection on MATH-500. \Cref{fig:scaling-training-data-llama,fig:scaling-training-data-qwen} in \Cref{app:scaling-training-data} show a performance boost from training on the 65K examples compared to only 1K. This suggests that \model can utilize more training data when available.

\subsection{Effect of Verification CoT Length on PRM Quality}

We study whether the length of verification chains of thought affects the quality of the resulting generative verifier.  
Specifically, we compare ThinkPRM trained on the full, long synthetic CoTs with a variant trained on short, compressed versions of the same 1K CoTs.  
To obtain the short CoTs, we instruct \texttt{gpt-4o-mini} to rewrite each original CoT into a concise version that preserves only the essential reasoning.  
We then train R1-Qwen-1.5B and R1-Qwen-14B on these short CoTs and evaluate verification F1 on ProcessBench.  
Table~\ref{tab:cot-length} reports the comparison.

\begin{table}[h!]
\centering
\small
\begin{tabular}{lcccc}
\toprule
& \multicolumn{2}{c}{Long CoT (\model)} & \multicolumn{2}{c}{Short CoT} \\
\cmidrule(lr){2-3} \cmidrule(lr){4-5}
\textbf{Model} & OlympiadBench & OmniMath & OlympiadBench & OmniMath \\
\midrule
R1-Qwen-1.5B & 87.3 & 75.7 & 64.8 & 66.7 \\
R1-Qwen-14B  & 87.3 & 85.7 & 55.3 & 60.8 \\
\bottomrule
\end{tabular}
\caption{Verification F1 when training R1 models on long versus short CoTs.}
\label{tab:cot-length}
\end{table}

The substantial performance drop when training on short CoTs emphasizes how \model benefits from extended reasoning. Since verification is a complex task, throwing more reasoning effort  at it via thinking improves performance. These results support the value of using long verification CoTs for training.

\subsection{Reasoning traces without clear step boundaries}
\label{sec:qwen3-ver}

So far, we have used \model to verify short CoTs with clear steps delimiters. Here, we investigate whether \model can still verify long CoTs that involve extended reasoning, backtracking, and self-correction. As a generator, we use Qwen3-1.7B \citep{yang2025qwen3} with thinking mode. Although \model was only trained on short solutions from PRM800K, it can still verify long CoTs and outperforms the baselines as shown in \Cref{fig:qwen3bon-computematchedmath-computematched-gpqa} left. Inspecting \model's outputs, we found that it extracts and verifies individual steps embedded in the long CoT---an example is in \Cref{fig:qwen3-cot}.

\subsection{Compute-matched comparison to self-consistency}
\label{sec:compute-matched}

\begin{figure}
    \centering
    \includegraphics[width=0.32\linewidth]{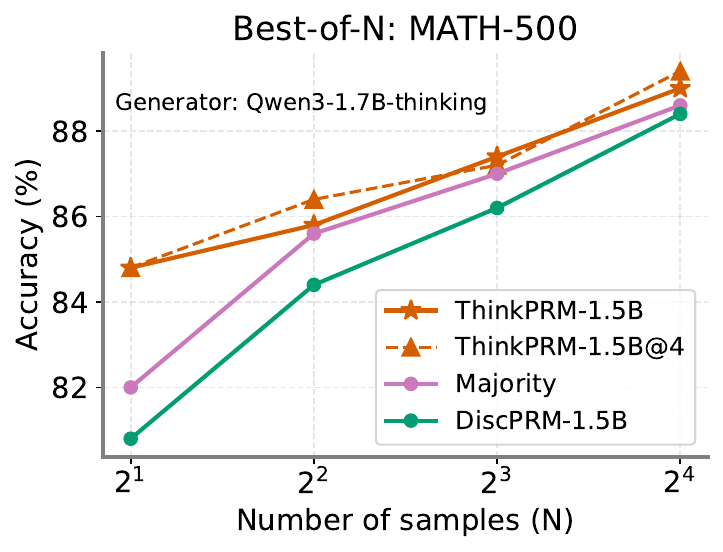}
    \includegraphics[width=0.32\linewidth]{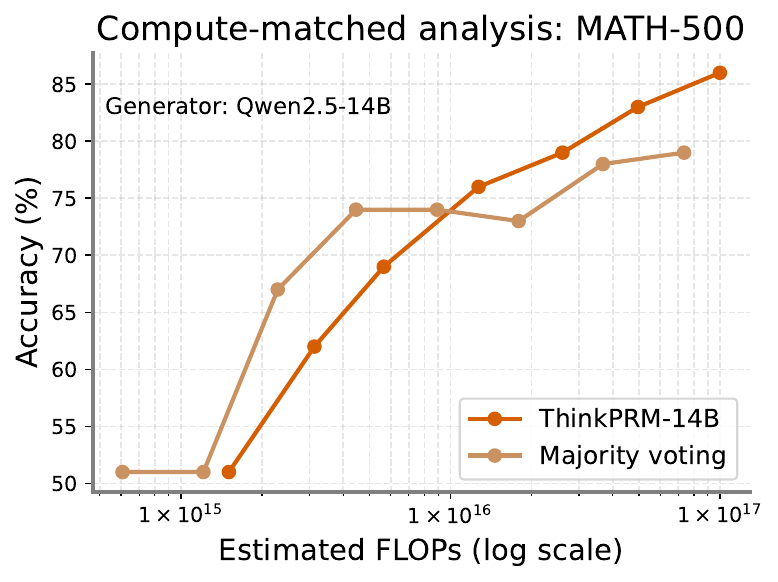}
    \includegraphics[width=0.33\linewidth]{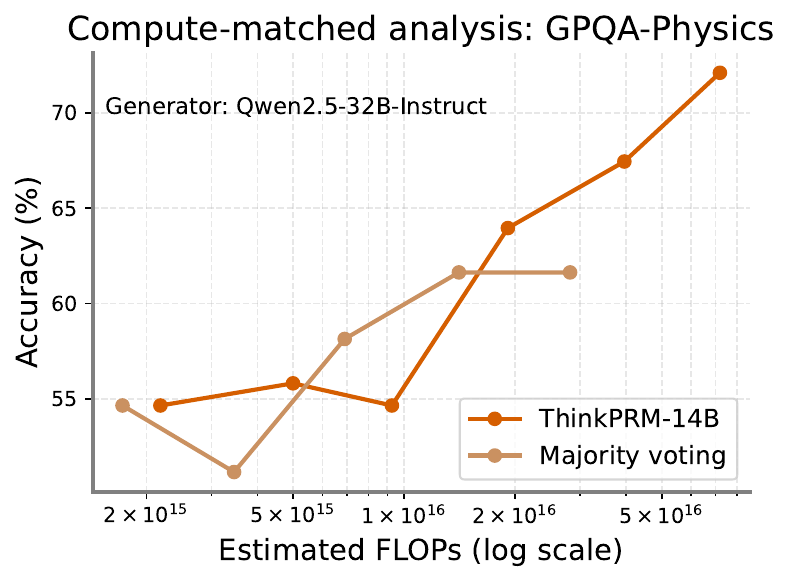}
    \caption{\textbf{Left: }Best-of-N with Qwen3-1.7B on the full MATH-500 test set, showing how \model generalizes well to verifying long reasoning traces. \textbf{Mid and Right:} Compute-matched comparison between best-of-N with \model and self-consistency or majority vote.}
    \label{fig:qwen3bon-computematchedmath-computematched-gpqa}
\end{figure}

Under a fixed test-time compute budget for best-of-N, how does \model compare to simply sampling more solutions from the generator and applying majority voting? To investigate this, we conduct a compute-matched analysis on MATH-500 and GPQA-Physics. \Cref{fig:qwen3bon-computematchedmath-computematched-gpqa} mid and right plot solution accuracy as a function of sampling FLOPs for MATH-500 and GPQA-physics. At low sampling budgets, best-of-N with \model performs comparably to self-consistency, but as the compute budget increases, \model has a clear advantage. These findings agree with recent work on outcome reward models \citep{singhi2025solve}.

\subsection{\model with Monte Carlo step labels}
\label{sec:automatic-labels}
To train \model, we have relied on manual step labels from PRM800K. Since automatic labels e.g., via Monte Carlo rollouts \cite{luo2024improve} are cheaper, we validate whether we can train \model using automatic labels. We train \model-1.5B using 1K synthetic chains based on labels from Math-shepherd dataset \citep{math-shepherd}. Performance on ProcessBench is shown in \Cref{tab:automatic-labels}, where training \model with automatic labels yields very comparable performance to training with manual labels, showing that our training pipeline is agnostic to step-labeling strategy.

\subsection{\model helps with difficult reasoning problems}
\label{sec:difficulty-analysis}
\model's reasoning ability should enable it to tackle verification of hard problems. To check if this is the case, we analyze performance of \model vs. DiscPRM in light of problem difficulty over MATH-500 and GPQA-physics (how we estimate difficulty for GPQA-Physics is explained in \Cref{app:difficulty}), shown in \Cref{fig:difficulty-analysis}. The generators here are Qwen-2.5-14B for MATH-500 and Qwen-2.5-32B-Instruct for GPQA-Physics. Primarily, \model improves reasoning on the \textit{difficult} problems (levels 3, 4, 5 in MATH-500 and 2, 3, 4 in GPQA-Physics) substantially more than DiscPRM.

\subsection{Filtering based on process vs. outcome labels}
\label{sec:filtering-ablation}
In \Cref{sec:finetuning}, we describe our process-based filtering strategy, which selects verification CoTs based on agreement between generated step-level decisions and gold process labels. To validate its effectiveness, we compare it to outcome-based filtering, as in GenRM~\citep{zheng2024processbench}, which retains chains solely based on final answer correctness—keeping a CoT if its final answer is correct and the final step is \texttt{\textbackslash boxed\{correct\}}, or if the answer is incorrect and the final step is \texttt{\textbackslash boxed\{incorrect\}}, thereby ignoring intermediate step labels. We obtain 65K and 128K CoTs using process- and outcome-based filtering, respectively. \Cref{fig:process-vs-outcome} shows that finetuning R1-Distill-Qwen-1.5B on process-filtered data yields significantly better verification performance, despite using fewer examples, which reflects the importance of our process-based filtering in training strong PRMs.

\subsection{Limitations of Generative PRMs}
\label{sec:limitations}
While generative PRMs are more powerful and data-efficient than their discriminative counterparts, they come with some limitations that we highlight as avenues for future work. First, overconfidence is a known issue in LLMs~\citep{liu2023litcab,stechly2023gpt,zhou2024relying} and, in the case of PRMs, it can cause the predicted PRM scores to cluster near extremes: close to either 0 or 1. One reason is that we are using probabilities of certain tokens such as ``yes'' or ``no'', which by nature will be either very high or very low. Future work should explore more reliable techniques to extract calibrated scores from generative reward models. Another limitation is due to autoregressive nature of LLMs, leading them to prematurely commit to an earlier judgment. For example, we observe a phenomenon we term \textit{step label interference}, where verification errors for earlier steps impact verification of later steps. 
For example, we noted that if the PRM judges a particular step as incorrect, it becomes more likely to label subsequent steps as incorrect even if it is not.
Lastly, generating a verification CoT introduces extra overhead compared to discriminative PRMs, but we argue that the performance gains offered by generative PRMs justify this extra cost.

\section*{Conclusion}
We introduced \model, a generative process reward model trained with minimal synthetic supervision for scalable step-by-step verification. With just 8K process labels, \model significantly outperforms LLM-as-a-judge and even surpasses discriminative PRMs trained on orders of magnitude more data. These results highlight the benefits of generative PRMs in interpretability, scalability, and data efficiency, and demonstrate their potential to scale verification compute for complex reasoning tasks in math and science.

\bibliography{ref}
\bibliographystyle{colm2025_conference}

\appendix
\crefalias{section}{appendix}
\newpage

\section{Training data}

\subsection{Sampling}
\label{app:data-collection}

\Cref{fig:data-collection} shows the prompt used with QwQ-32B-Preview to sample verification chains for training. We use the problem-prefix pairs from PRM800K train split \citep{lightman2023let}, which is based on MATH \citep{hendrycks2021measuring}. We sample 4 verification chains for each prefix in the dataset with temperature $T=0.1$ until we have 1K verification CoTs that (1) follow the desired format i.e., the model generated an extractable label for every step in the prefix and (2) the verbalized step labels match the gold labels in the PRM800K dataset annotation, (3) at most 4096 tokens long, and are (4) roughly balanced with respect to final answer correctness. 

Roughly 20\% of the initial samples satisfied criteria 1 and 2 in our case, which means that about 5K samples in total are needed to obtain 1K filtered samples. \Cref{tab:data-statistics} shows the statistics of the final dataset. We also show an example from the 1K training CoTs in \Cref{fig:train-example}.

\begin{figure}
\footnotesize
\begin{tcolorbox}[colback=blue!0!white, colframe=orange!75!black, title=Synthetic training sample,fonttitle=\bfseries]
\textbf{Problem:} Find the area in square feet of a square with a perimeter of 32ft.

\textbf{Solution:}\\
Step 1: A square has four equal sides, so the perimeter is four times the length of one side.\\
Step 2: I can write an equation to relate the perimeter and the side length: 4s = 32, where s is the side length in feet.\\
Step 3: To solve for s, I can divide both sides by 4: s = 32/4 = 8.\\
Step 4: Now I know the side length of the square is 8 feet.\\
Step 5: The area of a square is the side length squared, so the area is s² = 8² = 64.\\
Step 6: The area is in square feet, so the units are ft².\\
Step 7: The answer is 64 ft² \\ 

\textcolor{orange}{\texttt{<think>}}\\
Alright, let's verify step by step the solution provided for finding the area of a square with a perimeter of 32 feet.

\vspace{0.3em}
\textbf{Step 1:} ``A square has four equal sides, so the perimeter is four times the length of one side."\\
Critique: This statement is accurate. A square indeed has four sides of equal length, and the perimeter is the sum of all sides. Therefore, perimeter (P) equals 4 times the side length (s), or P = 4s.\\
The step is \textbf{\textbackslash\textbackslash{boxed}\{correct\}}

\vspace{0.3em}
\textbf{Step 2:} ``I can write an equation to relate the perimeter and the side length: 4s = 32, where s is the side length in feet."\\
Critique: Given that the perimeter is 32 feet, setting up the equation 4s = 32 is correct. It properly represents the relationship between the perimeter and the side length of the square.\\
The step is \textbf{\textbackslash\textbackslash{boxed}\{correct\}}

\vspace{0.3em}
\textbf{Step 3:} ``To solve for s, I can divide both sides by 4: s = 32/4 = 8."\\
Critique: This step correctly solves for s by dividing both sides of the equation by 4. The calculation 32 divided by 4 equals 8 is accurate.\\
The step is \textbf{\textbackslash\textbackslash{boxed}\{correct\}}

\vspace{0.3em}
\textbf{Step 4:} ``Now I know the side length of the square is 8 feet."\\
Critique: Based on the previous step, s = 8 feet, which is correctly stated here.\\
The step is \textbf{\textbackslash\textbackslash{boxed}\{correct\}}

\vspace{0.3em}
\textbf{Step 5:} ``The area of a square is the side length squared, so the area is s² = 8² = 64."\\
Critique: The formula for the area of a square is indeed s squared. Calculating 8 squared equals 64 is correct.\\
The step is \textbf{\textbackslash\textbackslash{boxed}\{correct\}}

\vspace{0.3em}
\textbf{Step 6:} ``The area is in square feet, so the units are ft²."\\
Critique: Since the side length is in feet, the area will be in square feet (ft²). This is the correct unit for area.\\
The step is \textbf{\textbackslash\textbackslash{boxed}\{correct\}}

\vspace{0.3em}
\textbf{Step 7:} ``The answer is 64 ft²"\\
Critique: This is the final answer, correctly stating the area as 64 square feet.\\
The step is \textbf{\textbackslash\textbackslash{boxed}\{correct\}}

\vspace{0.3em}
\textcolor{orange}{\texttt{</think>}}\\
\textbf{Is the solution correct?} Yes
\end{tcolorbox}
\caption{Training example from our sampled 1K synthetic chains.}
\label{fig:train-example}
\end{figure}

\subsection{Preprocessing}
From the sampled verification CoTs, we extract boxed predictions (e.g., \verb|\boxed{correct}| or \verb|\boxed{incorrect}|) for each reasoning step. Then we filter the data as follows:
    \begin{itemize}[leftmargin=*]
    \item CoTs with missing or malformed verification decisions are discarded.
    \item CoTs where verification decisions don't match ground truth labels are filtered out. Precisely, we require all step-by-step verification decisions to match the ground truth labels.
    \end{itemize}

Then we clean the data as follows:
    \begin{itemize}[leftmargin=*]
        \item We standardize boxed notation formats.
        \item Content after the last verification decision is removed.
        \item Special tokens (e.g., \verb|<think>| and \verb|</think>|) are added to designate verification reasoning.
    \end{itemize}

\begin{table}[h]
\centering
\caption{Statistics of the 1K finetuning dataset.}
\footnotesize
\label{tab:data-statistics}

\begin{tabular}{@{}ll@{}}
\toprule
\textbf{Correctness Distribution} &  \\
\# correct & 486 {\scriptsize(48.6\%)} \\
\# incorrect & 514 {\scriptsize(51.4\%)} \\
\midrule
\textbf{Step label distribution} &  \\
\# correct  & 7474 {\scriptsize(92.3\%)} \\
\# Incorrect & 625 {\scriptsize(7.7\%)} \\
\midrule
\textbf{Prefix statistics} &  \\
\# unique questions & 869 \\
average steps per prefix & 9.5 \\
min \# steps & 1 \\
max \# steps & 53 \\
\midrule
\textbf{Verification Chain Statistics} &  \\
Average verification chain length (tokens) & 1037.0 \\
Min verification chain length (tokens) & 207 \\
Max verification chain length (tokens) & 3669 \\
\bottomrule
\end{tabular}
\end{table}

\begin{figure}
\footnotesize
\begin{tcolorbox}[colback=blue!0!white, colframe=orange!75!black, title= Data generation prompt]
You are given a math problem and a proposed multiple-step solution (with a step on each line):\\

\textbf{[Math Problem]} \\
\{problem\} \\

\textbf{[Solution]} \\
\{solution\} \\

Review and critique the proposed solution steps and determine whether each step is correct. If the solution is incomplete, only critique the steps that are provided. Your output must be in the following format:\\

Let's verify step by step:\\
Step 1: \textless critique\textgreater ...The step is \textbackslash boxed\{correct/incorrect\} \\
Step 2: \textless critique\textgreater ...The step is \textbackslash boxed\{correct/incorrect\} \\
\ldots \\
Step n: \textless critique\textgreater ...The step is \textbackslash boxed\{correct/incorrect\} \\

Once you find an incorrect step, you should stop since you don't need to analyze the remaining steps.
\end{tcolorbox}
\caption{Prompt used to generate synthetic verification chains as described in \Cref{sec:finetuning}.}
\label{fig:prompt-data-generation}
\end{figure}
\newpage

\section{LLM-as-a-Judge Limitations}
\label{app:issues}
We show examples of failures from QwQ-32B-Preview and Deepseek-R1-Distill-Qwen-14B when used for LLM-as-judge verification in \Cref{fig:repetition-example-2,fig:incorrect-judgement}. 
\begin{figure}[]
\scriptsize
\begin{tcolorbox}[colback=red!10!white,colframe=red!30!black,title=QwQ-32B-Preview stuck in a loop]
Okay, so I need to analyze each step of the proposed solution to the math problem. The problem is: If \( f(x) = \frac{3x - 2}{x - 2} \), what is the value of \( f(-2) + f(-1) + f(0) \)? The proposed solution is given, and I have to check if each step is correct.\\

First, let me read through the problem and the solution carefully.\\

The function is \( f(x) = \frac{3x - 2}{x - 2} \). They want the sum of \( f(-2) + f(-1) + f(0) \).\\

Looking at the solution steps:\\

\textbf{Step 0}: They plug in \( x = -2 \) into the function and calculate \( \frac{3(-2) - 2}{-2 - 2} = \frac{-8}{-4} = 2 \).\\

\textbf{Step 1}: They mention that plugging \( x = -2 \) into the denominator gives zero, which would be division by zero.\\

\textbf{Step 2}: They generalize that we should avoid plugging in values that make the denominator zero.\\

\textbf{Step 3}: Similarly, plugging \( x = -1 \) into the denominator also gives zero.\\

\textbf{Step 4}: Plugging \( x = 0 \) into the denominator also results in division by zero.\\

\ldots\\

Wait no! The original function is given as \( f(x) = \frac{\text{numerator}}{\text{denominator}} = ? \)\\

No! The problem says: If \( f(x) = \frac{\text{numerator}}{\text{denominator}} = ? \)\\

No! The problem says: If \( f(x) = \frac{\text{numerator}}{\text{denominator}} = ? \)\\

No! The problem says: If \( f(x) = \frac{\text{numerator}}{\text{denominator}} = ? \)\\

No! The problem says: If \( f(x) = \frac{\text{numerator}}{\text{denominator}} = ? \)\\

No! The problem says: If \( f(x) = \frac{\text{numerator}}{\text{denominator}} = ? \)\\

No! The problem says: If \( f(x) = \frac{\text{numerator}}{\text{denominator}} = ? \)\\

No! The problem says: If \( f(x) = \frac{\text{numerator}}{\text{denominator}} = ? \)\\

No! The problem says: If \( f(x) = \frac{\text{numerator}}{\text{denominator}} = ? \)\\

No! The problem says: If \( f(x) = \frac{\text{numerator}}{\text{denominator}} = ? \)\\
... \\
\end{tcolorbox}
\label{tab:repetition-example}
\caption{Degeneration behavior from QwQ-32B-Preview with LLM-as-judge verification. Problem and solution are omitted for brevity.}

\end{figure}

\begin{figure}[]
\scriptsize
\begin{tcolorbox}[colback=red!10!white,colframe=red!30!black,title=Deepseek-R1-Distill-Qwen-14B stuck in a loop]

Okay, so I need to review and critique the solution step by step for the given math problem. The problem is to find the exact value of the continued fraction expression: 

1 + 1/(1 + 2/(1 + 1/(1 + 2/(1 + ...))))

The solution provided is split into 16 steps, each enclosed in a tag. I need to go through each step, analyze it, and see if there are any errors or if the logic holds. After that, I'll determine if the solution is correct by outputting "yes" or "no" in a box.

Let me start by understanding the problem. It's a continued fraction that alternates between 1 and 2 in the numerators. The expression is:

x = 1 + 1/(1 + 2/(1 + 1/(1 + 2/(1 + ...))))

So, the pattern repeats every two steps: 1, then 2, then 1, then 2, etc. The solution starts by denoting the expression as x and then tries to find a relationship by substituting the inner expression back into x.

Wait, the original expression is x = 1 + 1/(1 + 2/(1 + 1/(1 + 2/(1 + ...)))). So, the part after the first 1 is 1 + 2/(1 + 1/(1 + 2/(1 + ...))). But that part isn't x, because x starts with 1 + 1/(...). So, maybe the substitution isn't directly x. Let me think.

Wait, perhaps the inner part is similar but not exactly x. Let me denote the inner part as y. So, y = 1 + 2/(1 + 1/(1 + 2/(1 + ...))). But then, the expression for y would involve another similar structure. Hmm, maybe the solution is oversimplifying here. \\ 
... \\ 

Wait, actually, when you multiply both sides by (1 + 2/x), the left side is x*(1 + 2/x) = x + 2.

The right side is [1 + 1/(1 + 2/x)]*(1 + 2/x) = 1*(1 + 2/x) + [1/(1 + 2/x)]*(1 + 2/x) = (1 + 2/x) + 1 = 2 + 2/x.

So, the correct equation after multiplying both sides by (1 + 2/x) is:

x + 2 = 2 + 2/x.

But in step 6, they have:

x + 2 = 1 + 1/(1 + 2/x) + 2/x + 4/x²

Which seems to be incorrect. They seem to have distributed the multiplication incorrectly.

Wait, let me see. If you have x*(1 + 2/x) = [1 + 1/(1 + 2/x)]*(1 + 2/x) + [2/x*(1 + 2/x)]

But that doesn't make sense because when you multiply both sides by (1 + 2/x), you should only have:

Left side: x*(1 + 2/x)

Right side: [1 + 1/(1 + 2/x)]*(1 + 2/x)

Which simplifies to x + 2 = 2 + 2/x.

But in the solution, step 6 says:

x*(1 + 2/x) = 1 + 1/(1 + 2/x) + 2/x*(1 + 2/x)

Which seems to be adding an extra term, 2/x*(1 + 2/x), which is not correct. So, step 6 is incorrect.

Therefore, the solution has an error in step 6. Instead of correctly multiplying both sides, they incorrectly expanded the right-hand side, leading to an incorrect equation.

Because of this mistake, the subsequent steps are likely incorrect as well. So, the entire solution is flawed from step 6 onwards.

...

Wait, let me compute the continued fraction numerically to check.

Let me compute the continued fraction up to a certain depth.

Let me compute it step by step:

Start with the innermost part and work outwards.

... \\ 

So, it seems to converge to approximately 2.414, which is 1 + sqrt(2) $\approx$2.4142.

Wait, but according to my earlier solving, x should be sqrt(2) $\approx$1.414, but the numerical computation suggests it converges to approximately 2.414, which is 1 + sqrt(2). So, which one is correct?

Wait, perhaps I made a mistake in setting up the equation.

Wait, let's re-examine the original expression: \\ 

... \\ 

But when I tried to solve it earlier, I set x = 1 + 1/(1 + 2/x), which led to x = sqrt(2). But the numerical computation suggests it converges to 1 + sqrt(2). So, perhaps my initial substitution was incorrect.

Wait, perhaps I need to define it differently. Let me try to define the continued fraction properly.

Wait, let's solve the system:

From x = 1 + 1/y and y = 1 + 2/x.

Substitute y = 1 + 2/x into x = 1 + 1/y:

x = 1 + 1/(1 + 2/x)

Multiply both sides by (1 + 2/x): \\ 

... \\ 

Alternatively, perhaps I need to define y differently. Let me try to define y as the expression starting with 1 + 1/(...):

y = 1 + 1/(1 + 2/(1 + 1/(1 + 2/(1 + ...))))

But then, y = x, which brings us back to the original substitution.

... \\ 

Which is the same equation as before, leading to x = sqrt(2). But the numerical computation suggests it's approaching 1 + sqrt(2). So, perhaps my initial substitution is incorrect.

Alternatively, perhaps the continued fraction is structured such that after the first 1 + 1/(...), the denominator is another continued fraction similar to x but shifted.

...
\end{tcolorbox}
\caption{Degeneration behavior from Deepseek-R1-Distill-Qwen-14B with LLM-as-judge verification. Problem and solution are omitted for brevity.}
\label{fig:repetition-example-2}

\end{figure}
\begin{figure}[htbp]
\scriptsize
\begin{tcolorbox}[colback=blue!0!white, colframe=orange!75!black, title=Problem and solution from ProcessBench]

\textbf{Problem:} Let \( a, b \) be positive reals with \( a > b > \frac{1}{2}a \). Place two squares of side lengths \( a, b \) next to each other, such that the larger square has lower left corner at \( (0, 0) \) and the smaller square has lower left corner at \( (a, 0) \). Draw the line passing through \( (0, a) \) and \( (a + b, 0) \). The region in the two squares lying above the line has area 2013. If \( (a, b) \) is the unique pair maximizing \( a + b \), compute \( \frac{a}{b} \).

\vspace{1em}

\textbf{Solution:}

We begin by finding the equation of the line through \( (0, a) \) and \( (a + b, 0) \). The slope is:

\[
m = \frac{0 - a}{(a + b) - 0} = -\frac{a}{a + b}
\]

Using point-slope form from \( (0, a) \):

\[
y = a - \frac{a}{a + b}x
\]

\textbf{Area above the line in the larger square:}

At \( x = a \), the height is:

\[
y = a - \frac{a^2}{a + b} = \frac{ab}{a + b}
\]

So the area is a trapezoid:

\[
\text{Area}_1 = \frac{1}{2} \left( a + \frac{ab}{a + b} \right) a 
= \frac{a^3 + 2a^2b}{2(a + b)}
\]

\newpage

\textbf{Area above the line in the smaller square:}

This is a triangle with height \( \frac{ab}{a + b} \) and base \( b \):

\[
\text{Area}_2 = \frac{1}{2} \cdot \frac{ab}{a + b} \cdot b = \frac{ab^2}{2(a + b)}
\]

\textbf{Total Area:}

\[
\text{Total} = \text{Area}_1 + \text{Area}_2 = \frac{a^3 + 2a^2b + ab^2}{2(a + b)} 
= \frac{a(a^2 + 2ab + b^2)}{2(a + b)} = \frac{a(a + b)^2}{2(a + b)} = \frac{a(a + b)}{2}
\]

We are given:

\[
\frac{a(a + b)}{2} = 2013 \Rightarrow a(a + b) = 4026
\]

Express \( b \) in terms of \( a \):

\[
b = \frac{4026}{a} - a
\]

Apply the constraints \( a > b > \frac{1}{2}a \):

\begin{itemize}
  \item \( a > b \Rightarrow a > \frac{4026}{a} - a \Rightarrow a^2 > 2013 \Rightarrow a > \sqrt{2013} \)
  \item \( b > \frac{1}{2}a \Rightarrow \frac{4026}{a} - a > \frac{1}{2}a \Rightarrow a^2 < 2684 \Rightarrow a < \sqrt{2684} \)
\end{itemize}

\vspace{1em}

So \( \sqrt{2013} < a < \sqrt{2684} \). To maximize \( a + b = \frac{4026}{a} \), we minimize \( a \), so let \( a = \sqrt{2013} \). Then \( b = \sqrt{2013} \), satisfying the constraints.

\[
\frac{a}{b} = \boxed{2}
\]

\end{tcolorbox}
\caption{An example of problem and solution from OmniMath subset of ProcessBench.}
\label{tab:problem-prefix-for-incorrect-judgement}
\end{figure}

\begin{figure}[htbp]
\scriptsize
\begin{tcolorbox}[colback=red!10!white, colframe=red!30!black, title=Incorrect verification]

Okay, so I need to review and critique the given solution step by step. Let me go through each step carefully.

\textbf{Step 0:} The solution starts by identifying key points and properties. It mentions that \( J \) is the excenter opposite to \( A \), which is correct. \( A_1 \) is where the excircle touches \( BC \), and \( B_1, C_1 \) are where it touches the extensions of \( AC \) and \( AB \). Then, it says \( A_1B_1 \) is perpendicular to \( AB \) and intersects at \( D \). \( E \) is the foot of the perpendicular from \( C_1 \) to \( DJ \).

Hmm, this seems okay, but I wonder if they've correctly identified all the properties. Maybe they should have mentioned more about the excenter's properties, like the fact that it's the intersection of external angle bisectors. But maybe that's assumed knowledge.

\textbf{Step 1:} They note that since \( A_1B_1 \) is perpendicular to \( AB \), \( D \) is the projection of \( A_1 \) onto \( AB \). So, \( D \) lies on \( AB \) and \( A_1D \) is perpendicular to \( AB \).

Wait, is that necessarily true? If \( A_1B_1 \) is perpendicular to \( AB \), and they intersect at \( D \), then yes, \( D \) is the foot of the perpendicular from \( A_1 \) to \( AB \). That seems correct.

\textbf{Step 2:} They examine the excircle and the triangle, mentioning that the excircle is tangent to \( BC \) at \( A_1 \), and the tangents from \( A_1 \) to \( B_1 \) and \( C_1 \) form right angles with the radii. Then, \( DJ \) is a line through the excenter and the intersection point of the perpendiculars from \( A_1 \) and \( B_1 \) to \( AB \).

Hmm, I'm not sure about this part. The line \( DJ \) is through the excenter \( J \) and point \( D \), which is the foot from \( A_1 \). But does it involve perpendiculars from \( A_1 \) and \( B_1 \)? I think the solution might be conflating different properties here. The excenter is equidistant from all sides, but \( DJ \) is just a line from \( D \) to \( J \), not necessarily involving perpendiculars from both \( A_1 \) and \( B_1 \). Maybe this is a misstatement.

\textbf{Step 3:} They state that since \( E \) is the foot of the perpendicular from \( C_1 \) to \( DJ \), angle \( C_1ED \) is 90 degrees. Then, they argue that angles \( BEA_1 \) and \( AEB_1 \) are equal due to symmetry and properties of the cyclic quadrilateral.

Wait, why is \( E \) symmetrically placed? The solution says it's because of the properties of the excircle and right angles, but I'm not entirely convinced. Just because \( E \) is a foot of a perpendicular doesn't automatically make it symmetric with respect to \( A_1B_1 \). Maybe there's a better reason, like cyclic quadrilaterals or similar triangles, but this step feels a bit hand-wavy.

\textbf{Step 4:} They conclude both angles are 45 degrees because of symmetry and right angles, forming a cyclic quadrilateral. But I'm not sure if the cyclic quadrilateral part is clearly established. Also, 45 degrees seems specific—does the problem necessarily lead to that?

Wait, maybe I should draw a diagram to visualize. If \( A_1B_1 \) is perpendicular to \( AB \), and \( E \) is the foot from \( C_1 \) to \( DJ \), which is connected to the excenter. Hmm, maybe the angles are indeed 45 degrees because of the right angles and the properties of the excentral triangle, but I'm not entirely certain.

\textbf{Overall Thoughts:} The solution seems to have some correct steps but might be making assumptions without sufficient justification. For example, in Step 3, the symmetry argument isn't fully explained. Also, in Step 2, the description of \( DJ \) might not be accurate. The conclusion of 45 degrees might be correct, but the reasoning feels a bit rushed.

\textbf{Final Answer} After considering the steps, I think the solution might have some gaps in justification but arrives at the correct answer. So, I'll say yes, but with some reservations.

\boxed{yes}
\end{tcolorbox}
\caption{Incorrect verification by Deepseek-R1-Distill-Qwen-14B for the problem-solution pair in \Cref{tab:problem-prefix-for-incorrect-judgement}.}
\label{fig:incorrect-judgement}
\end{figure}

\section{Training details}

\subsection{Training \model}
\label{app:model-training}
We train our models on the filtered 1K chains using LoRA \citep{hu2022lora} with rank $r=32$ and $\alpha=16$. Specifically, we finetune Qwen-14B-Instruct and R1-Deepseek (14B) for 3 epochs, and QwQ-32B-Preview for 5 epochs using LoRA. We use an effective batch size of 16 and a fixed learning rate of $4 \times 10^{-4}$ without warmup is used. Training took 1.5 hours for the 14B models and 4.5 hours for QwQ-32B-Preview on a single A100 80GB GPU. Without particularly found QwQ to be hard to train with LoRA and still generates a relatively high percentage of invalid judgments after training. Full training of the model will likely resolve these issues but that would require more compute than we have.

The R1-Distill-Qwen\{1.5B,7B\} models use \textit{full} training with the following parameters. The 1.5B model We trained for 3 epochs with an effective batch size of 32, using a constant learning rate of $6 \times 10^{-5}$ without decay or warmup. We train both models using four RTX A6000 48GB GPU using data parallel. Training the 1.5B model on the 1K chains took about half an hour and the 7B model about two hours.

\subsection{Training Discriminative Verifiers}
\label{app:disc-training}
We train R1-Qwen-14B for 1 epoch over the entire PRM800K dataset using two A100 80GB GPUs with a batch size of 8 and a learning rate of $6 \times 10^{-5}$. We use a constant learning rate scheduler with no warmup. Following prior work \citep{math-shepherd,prmlessons} We train the model using binary cross-entropy to maximize the probability of the tokens ``+" and ``-" for correct and incorrect steps, respectively. The R1-Qwen-1.5B model is trained with the same infrastructure with a batch size of 64 and a learning rate of $1 \times 10^{-4}$ with a warm up ratio of 10\%.

\section{Results on ProcessBench before and after finetuning}
\Cref{tab:pb-results} shows the performance numbers of LLM-as-a-Judge and \model on ProcessBench.

\begin{table}[h!]
    \footnotesize
    \centering
    \caption{Average F1-score on OlympiadBench and OmniMath subsets of ProcessBench \citep{zheng2024processbench} comparing LLM-as-a-Judge to \model finetuned on 1K examples. Random baseline for OlympiadBench is 39.1\% and for OmniMath is 32.7\%. Percentage of bad outputs (repetitions, invalid label formatting, overthinking, etc.) are shown in \textcolor{red}{red}. LLM-as-a-judge with reasoning models suffer from issues that limits their utility as generative verifiers. }
    \label{tab:pb-results}
    \begin{tabular}{lcc|cc}
        \toprule
        \multirow{2}{*}{Model} & \multicolumn{2}{c}{LLM-as-a-Judge} & \multicolumn{2}{c}{\model} \\
        \cmidrule(lr){2-3} \cmidrule(lr){4-5}
        & OlympiadBench & OmniMath & OlympiadBench & OmniMath \\
        \midrule
        Random baseline & 39.1 & 32.7 & 39.1 & 32.7 \\ 
        \midrule
        R1-Qwen-1.5B & 5.0 \highlightred{51.4} & 5.4 \highlightred{55.1} & 76.3 \highlightred{1.4} & 75.7 \highlightred{2.4} \\
        R1-Qwen-7B & 44.8 \highlightred{18.2} & 45.7 \highlightred{20.9} & 73.4 \highlightred{1.1}  & 74.0 \highlightred{1.4} \\
        R1-Qwen-14B & \textbf{72.8} \highlightred{13.3} & \textbf{67.8} \highlightred{18.6} & \textbf{87.3} \highlightred{2.3} & \textbf{85.7} \highlightred{2.3} \\
        QwQ-32B-preview & 50.6 \highlightred{7.9} & 55.5 \highlightred{10.9} & 73.1 \highlightred{15.1} & 73.2 \highlightred{7.9} \\
        \bottomrule
    \end{tabular}
        \label{tab:pb-results-detailed}

\end{table}

\section{Evaluation details}
\label{app:eval-details}
This section includes exact details on the test-time scaling shown in \Cref{sec:results}

\subsection{Predicting verification labels}
Following prior work \citep{snell2024scaling,beeching2024scalingtesttimecompute}, we aggregate scores from DiscPRM by using the score of the \textit{last} step. For \model, we first prompt the model to generate the verification chain up to a maximum of 8192 tokens, then we force decode the string \textit{``Is the solution correct?''} and use $\frac{P(\text{``yes"})}{P(\text{``yes"}) + P(\text{``no"})}$ as the solution score.

\subsection{Best-of-N selection}
We sample solutions using a temperature of $T=0.8$ for Llama-3.2-3B-Instruct and $T=0.4$ for Qwen-2.5-14B. We instruct all models to think step by step and put the final answer in \verb|\boxed{}|.  All our Best-of-N experiments use weighted majority voting, which scores final answers based on the sum of the verifier scores of their solutions \citep{uesato2022solving,wu2024inference,sun2024easy}except for our experiments on AIME '24, where we use the verifier score directly to rank the solution, as we found this to perform better for all verifiers. 

\subsection{Verifier-guided beam search}
Under verifier-guided beam  search, we sample candidate next steps and score them with the process verifier, then selects top-$K$ out of these to further expand and so on. Our implementation is based on \citep{snell2024scaling,beeching2024scalingtesttimecompute}, which maintains $N$ beams in total, and samples $M$ candidate next steps per beam. We set $M=4$ for all experiments and run search for a maximum of 20 steps per beam. To sample next steps, we use $T=0.6$ and use double newlines as the step delimiter. 

\subsection{Sequential scaling of verifier compute}
\label{sec:bf}

We achieve budget forcing \citep{muennighoff2025s1} by triggering the model to think again for $R$ rounds, where each round uses a unique trigger phrase that incites the model to revisit or double-check its earlier verification. We use different trigger phrases for each round since we found that using the same phrase causes the model to repeat what it did in the last round. 

We do a maximum of $R=4$ thinking rounds, and use the phrases ``Let me double check", ``Let's verify again", and  ``Did I miss something?'', for rounds 2, 3, and 4 respectively. We do not investigate deeply into optimizing the trigger phrase, but we note that performance may depend on these and we use the same phrases for both \model and LLM-as-a-judge to ensure fair comparison. 

\subsection{MATH-500 test examples}
\label{app:math-100}
\begin{figure}[h!]
  \centering
  \begin{floatrow}
    \ffigbox[\FBwidth]
      {\includegraphics[width=0.95\linewidth]{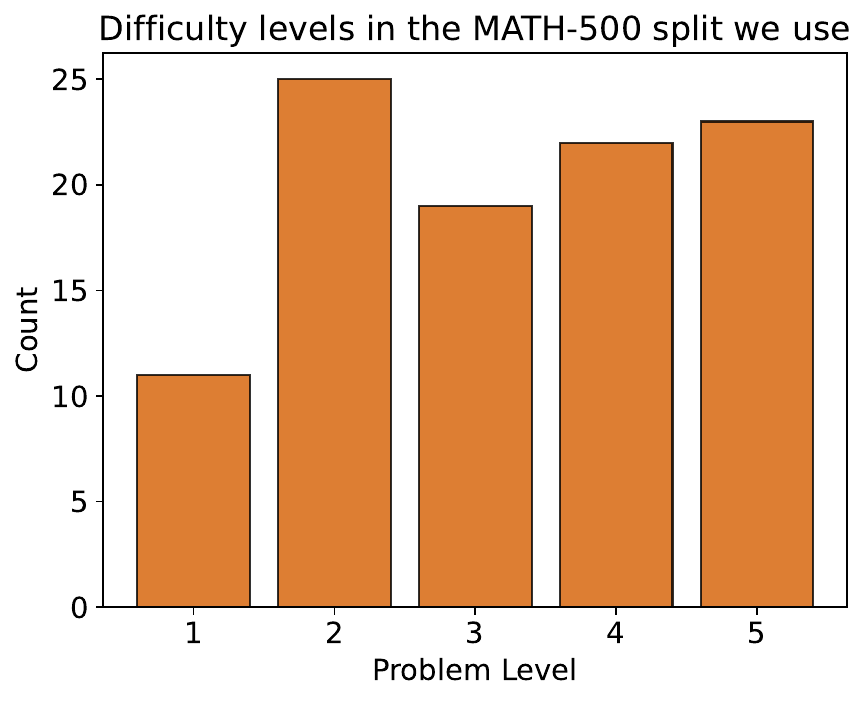}}
      {\caption{Histogram of difficulty levels in our 100-problem subset from MATH-500.}\label{fig:math-100-difficulty}}
    \ffigbox[\FBwidth]
      {\includegraphics[width=0.94\linewidth]{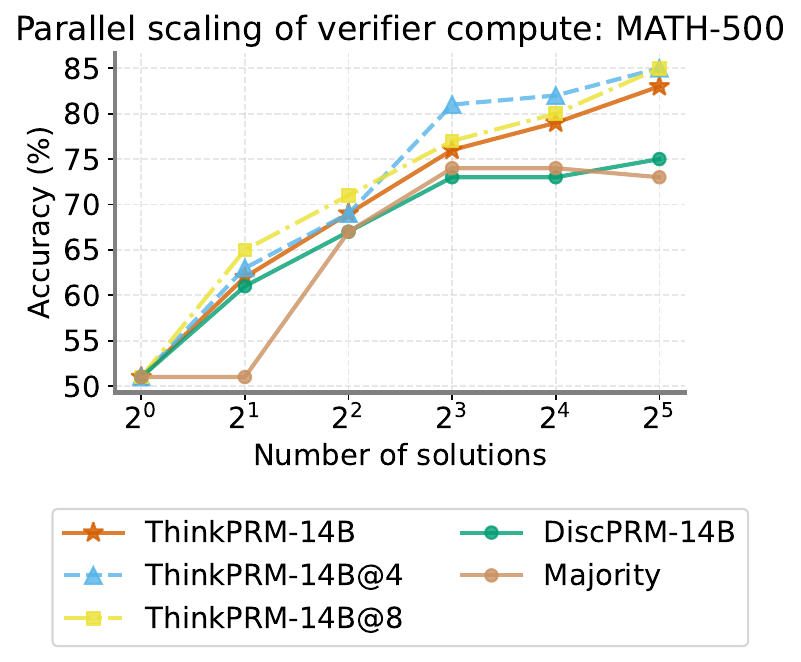}}
      {\caption{Scaling of verifier compute by parallel sampling of multiple verification CoTs and averaging their scores. Parallel scaling (\model-14B@4 and \model-14B@8) further boosts performance curve compared to scoring based on a single CoT (\model-14B).}\label{fig:parallel-scaling-math}}
  \end{floatrow}
\end{figure}

As running on all 500 examples from MATH-500 will require a lot of compute, we run all our experiments on 100 randomly sampled subsets from MATH-500 \citep{hendrycks2021measuring}. We pick the 100 problems such that they cover different difficulty levels, as shown in \Cref{fig:math-100-difficulty}.

\subsection{Additional results on scaling verifier compute}
\label{app:scaling-verifier-compute}

\begin{figure}[h!]
  \centering
  \begin{floatrow}
      \includegraphics[width=0.46\linewidth]{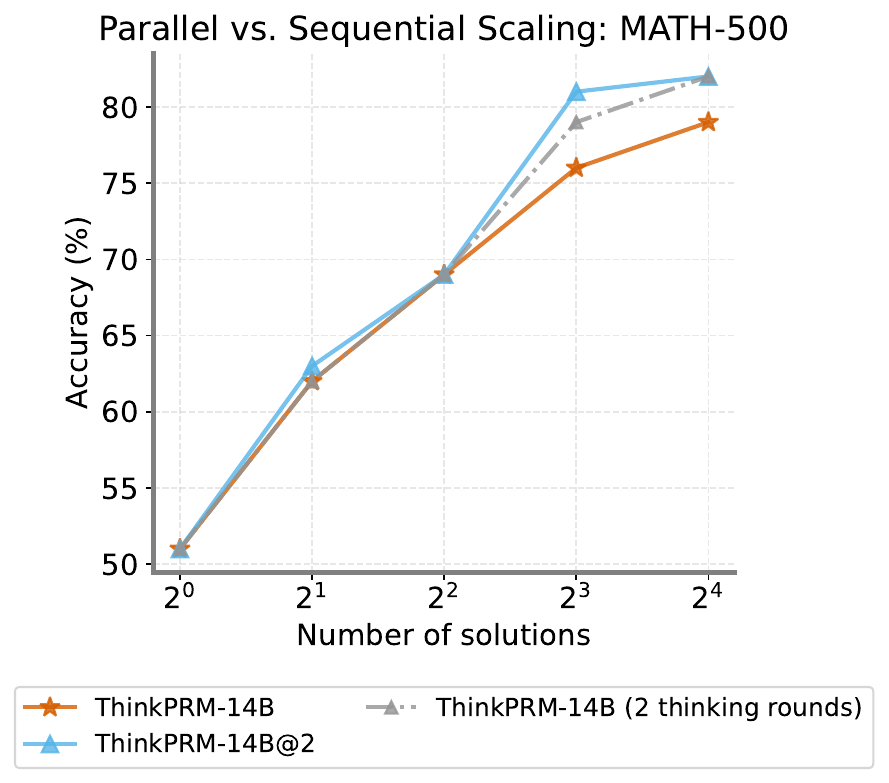}
      \includegraphics[width=0.46\linewidth]{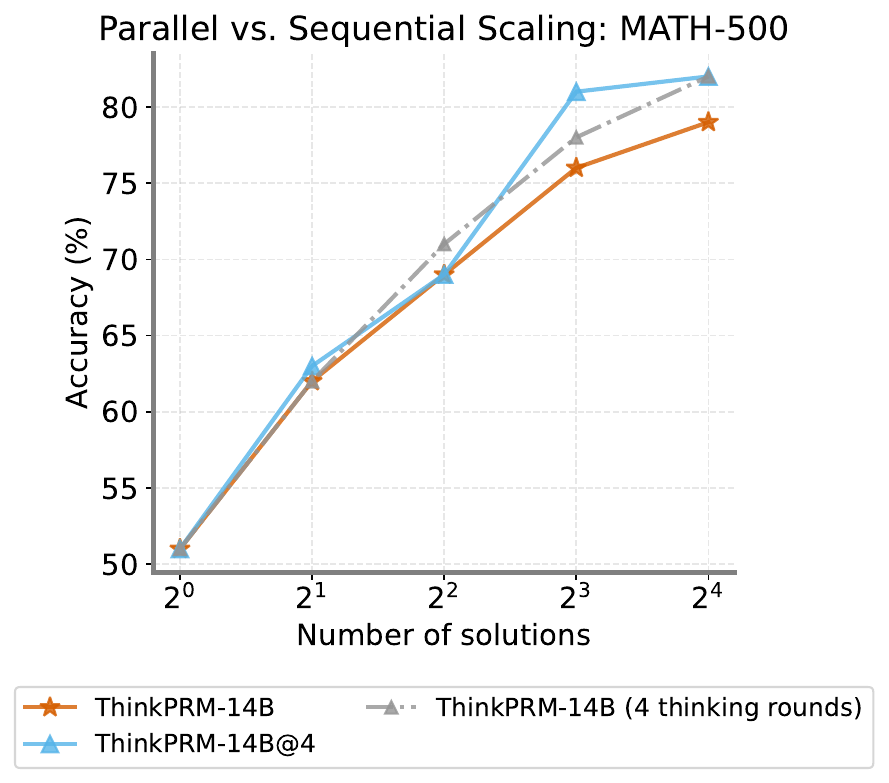}
      \caption{Parallel vs. sequential scaling of \model compute under the same generation budget with Qwen-2.5-14B generator. Parallel scaling (model@K) is done by independently sampling $K$ verification CoTs and aggregating their scores. Sequential scaling is done by prompting the model $K$ times to revise its own verification for $K$ thinking rounds. Both setups generate up until 8192 tokens per generation. We do not observe a clear winner although parallel scaling seems slightly better especially at larger sampling budgets.}
      \label{fig:parallel-vs-sequential}
  \end{floatrow}
\end{figure}

\Cref{fig:parallel-scaling-math} shows results of \model-14B of parallel scaling verifier compute by sampling $K=4$ and $K=8$ CoTs with temperature $T=0.6$ and aggregating their scores. Parallel scaling indeed lifts up the accuracy curve of \model-14B compared to standard $K=1$ with greedy decoding. However, performance plateaus rather quickly and $K=8$ remains comparable to $K=4$, while slightly better at smaller sampling budgets. \Cref{fig:parallel-vs-sequential} compares parallel to sequential scaling under the same token budget. While there is no clear winner, parallel scaling seems to perform slightly better at best-of-8.

\subsection{Scaling training data of \model}
\label{app:scaling-training-data}
\begin{figure}[h!]
  \centering
  \begin{floatrow}
    \ffigbox[\FBwidth]
      {\includegraphics[width=0.95\linewidth]{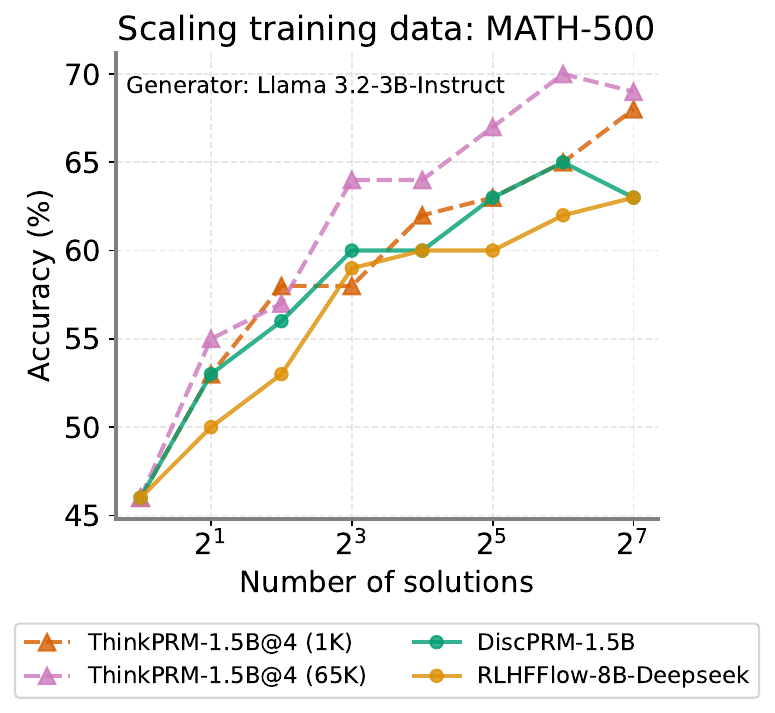}}
      {\caption{Best-of-N results with \model-1.5B comparing the version trained on 1K examples (used throughout the paper) and a version trained on 65K examples. \model benefits from training on more synthetic data as the performance can further improve with more training. }\label{fig:scaling-training-data-llama}}
    \ffigbox[\FBwidth]
      {\includegraphics[width=0.94\linewidth]{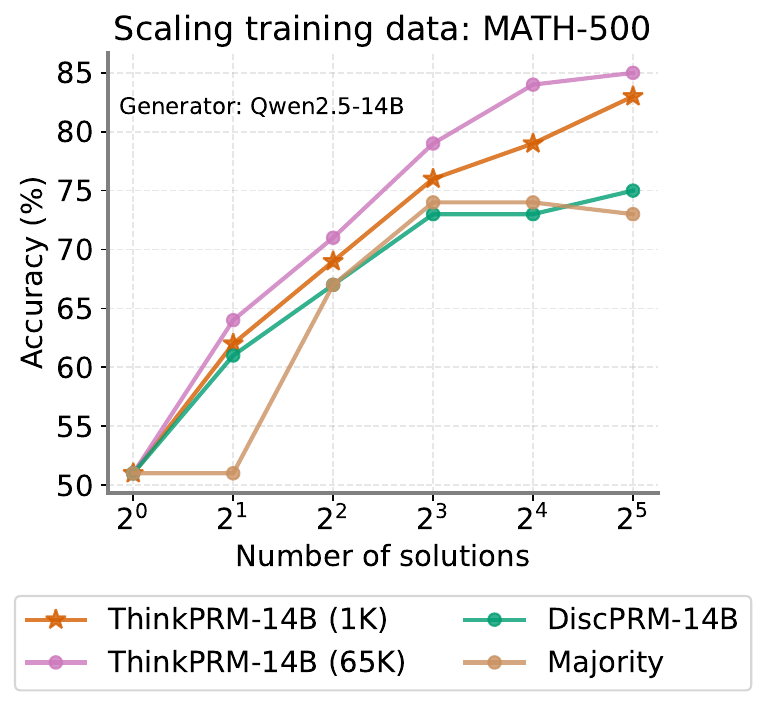}}
      {\caption{Best-of-N results with \model-14B comparing the version trained on 1K examples (used throughout the paper) and a version trained on 65K examples. \model benefits from training on more synthetic data as the performance can further improve with more training.}\label{fig:scaling-training-data-qwen}}
  \end{floatrow}
\end{figure}

Here, we show results when training \model-14B and \model-1.5B using synthetic data from all PRM800K. The goal is to show that \model can still benefit from training on more synthetic data. Here, we train both R1-Distill-Qwen-1.5B and R1-Distill-Qwen-14B on a total of 65K verification CoTs we obtained by sampling and filtering as explained in \Cref{sec:rq2}. \Cref{fig:scaling-training-data-llama,fig:scaling-training-data-qwen} show best-of-N performance with \model-1.5B and \model-14B respectively when trained on 65K and compares it to training on 1K examples. Interestingly, \model benefits from additional training, and can further improve the accuracy curve compared to the 1K-trained version on MATH-500. We note, however, that while training on more math data boosts performance on MATH-500, we observe some performance drop on out-of-domain tasks due to the distribution shift.

\subsection{Results with automatic labels}
\Cref{tab:automatic-labels} shows performance when filtering training data based on manual labels (PRM800K) vs automatic labels (Math-Shepherd) \citep{math-shepherd}. \model still performs well even with automatic labels, and comparably to manual labels.

\begin{table}[h!]
\centering
\begin{tabular}{@{}lcc@{}}
\toprule
\textbf{Model} & \textbf{OlympiadBench} & \textbf{OmniMath} \\
\midrule
ThinkPRM-1.5B (PRM800K)        & 76.3 & 75.7 \\
ThinkPRM-1.5B (Math-shepherd)  & 75.8 & 76.5 \\
\bottomrule
\end{tabular}
\caption{Comparison of ThinkPRM-1.5B trained on PRM800K vs Math-shepherd step labels.}
\label{tab:automatic-labels}
\end{table}

\subsection{Verifier performance in terms of problem difficulty}
\label{app:difficulty}

\begin{figure}[h!]
    \centering
    \includegraphics[width=0.85\linewidth]{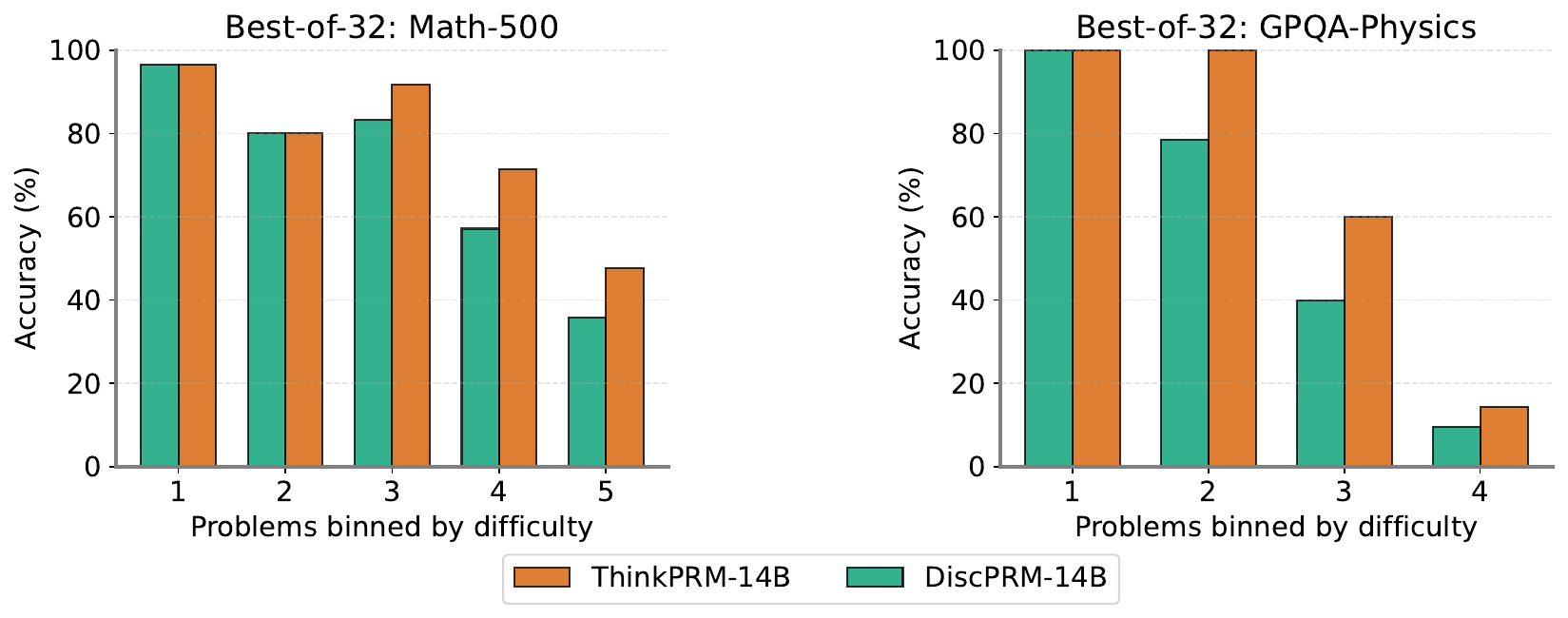}
    \caption{\model helps with challenging reasoning problems compared to DiscPRM. The generator model here is Qwen-2.5-14B for MATH-500 and Qwen-2.5-32B-Instruct for GPQA. }
    \label{fig:difficulty-analysis}
\end{figure}

We the difficulty  We do not estimate the difficulty over MATH problem since each problem in MATH is annotated based on 1 of 5 difficulty levels. For GPQA-Physics problems, we first compute the pass@1 rate of Qwen2.5-32B-Instruct for every problem by sampling 32 solutions for each problem. Then, we bin the pass@1 rates into four different bins and use the bin assignment for each problem as the difficulty level. As discussed in \Cref{sec:difficulty-analysis}, \Cref{fig:difficulty-analysis} show how \model helps more with verifying hard problems compared to DiscPRM.

\begin{table}[h!]
\centering
\footnotesize
\renewcommand{\arraystretch}{1.2} %
\begin{tabular}{llcc}
\toprule
\textbf{} & \textbf{Prompt} & \textbf{OlympiadBench} & \textbf{OmniMath} \\
\midrule
\multirow{3}{*}{QwQ-32B} 
 & Single Yes/No & 50.6 \highlightred{7.9} & 55.5 \highlightred{3.2} \\
 & Bad-step-index & 65.4 \highlightred{18.3} & 69.0 \highlightred{16.8} \\
 & Decision-for-each-step & 40.1 \highlightred{6.4} & 36.6 \highlightred{9.2} \\
\midrule
\multirow{3}{*}{R1-Qwen-14B} 
 & Single Yes/No & 72.8 \highlightred{13.3} & 67.8 \highlightred{18.6} \\
 & Bad step index & 77.7 \highlightred{17.0} & 76.6 \highlightred{17.9} \\
 & Decision-for-each-step  & 73.3 \highlightred{14.6} & 71.4 \highlightred{14.7} \\
\bottomrule
\end{tabular}
\caption{F1-score on OlympiadBench and OmniMath subsets from ProcessBench with different prompts. Ratio of invalid judgments is shown in \textcolor{red}{red}.}
\label{tab:different-prompts}
\end{table}

\begin{figure}[ht]
\scriptsize
\begin{tcolorbox}[colback=blue!1!white, colframe=orange!75!black, title=Example problem-solution pair from ProcessBench]

\textbf{Problem:} \\
In quadrilateral $A B C D$, $\mathrm{m} \angle B + \mathrm{m} \angle D = 270^{\circ}$. The circumcircle of $\triangle A B D$ intersects $\overline{C D}$ at point $E$, distinct from $D$. Given that $B C = 4$, $C E = 5$, and $D E = 7$, compute the diameter of the circumcircle of $\triangle A B D$. \\
\hrule\vspace{5pt}

\textbf{Solution:} \\ 

\textbf{Step 0:} To solve this problem, we will use several properties of geometry, particularly related to circles and quadrilaterals. Let's break down the steps: First, we understand the given information. We have a quadrilateral $A B C D$ where $\angle B + \angle D = 270^\circ$. The circumcircle of $\triangle A B D$ intersects $\overline{C D}$ at point $E$, distinct from $D$. Additionally, $B C = 4$, $C E = 5$, and $D E = 7$. \\

\textbf{Step 1:} Next, we apply properties of cyclic quadrilaterals and circles. Since $E$ lies on the circumcircle of $\triangle A B D$, $E$ is also part of the circumcircle of $\square A B C D$ (assuming $A B C D$ is cyclic, which will be justified later). In a cyclic quadrilateral, opposite angles sum up to $180^\circ$. Given $\angle B + \angle D = 270^\circ$, it suggests an error in our initial understanding or misinterpretation, since angles cannot sum to more than $360^\circ$ in a quadrilateral. However, if we consider the context, let's assume a correction was intended for the sake of explanation: Correct interpretation: If we adjust for logical consistency, perhaps the intention was to imply that one of the angles at $B$ or $D$ is $90^\circ$ (since this would fit the property of cyclic quadrilaterals better, aligning with the geometry principles involved). For this explanation, let's proceed with an assumption that either $\angle B$ or $\angle D$ is $90^\circ$, aiming for a coherent solution path. \\

\textbf{Step 2:} We then use the Power of a Point Theorem. The Power of a Point Theorem states that for a point $P$ outside a circle, the product of the lengths of the two segments from $P$ to the circle equals the product of the lengths of the other two segments when considering tangents or secants from that point. However, in this case, $E$ is not outside but on the circumference of the circle, and we apply it differently: Given $B C = 4$ and $C E = 5$, $D E = 7$, and knowing $E$ is part of the circumcircle, we can apply the property of intersecting chords in a circle, which states that the product of the lengths of the segments of one chord is equal to the product of the lengths of the segments of the other chord. Thus, in this scenario, we consider $B C$ and $C D$ as intersecting chords, where $C D = B C + C E = 4 + 5 = 9$. 

\[
B C \cdot C D = C E \cdot D E
\]

Substituting the known values:

\[
4 \cdot 9 = 5 \cdot D E
\]

\[
36 = 5 \cdot D E
\]

Thus, 

\[
D E = \frac{36}{5} = 7.2.
\] \\

\textbf{Step 3:} This discrepancy suggests a reevaluation of the given values or a misunderstanding in their application within the context provided. However, for the purpose of demonstrating the method: \\

\textbf{Step 4:} Finally, we find the diameter of the circumcircle of $\triangle A B D$. The key to finding the diameter involves recognizing that the circumcircle's radius or diameter is crucial for calculating distances involving points $A$, $B$, and $D$. However, without specific details about how $E$ relates to these points geometrically beyond being on the circle, or additional information such as the length of another segment or angle measures, directly calculating the diameter becomes complex. \\

\textbf{Step 5:} The question's focus on $E$ and its relation to $B C$, $C E$, and $D E$ might suggest looking at the properties of chords in a circle or applying trigonometric relationships based on the angles given initially. However, due to the discrepancy noted and the complexity introduced by the misinterpretation of the angle condition, a direct calculation for the diameter of the circumcircle of $\triangle A B D$ isn't straightforward with the information provided. \\

\textbf{Step 6:} Given the complexities and discrepancies in the original problem statement, particularly with the angles summing to $270^\circ$ in a quadrilateral, the solution path becomes less clear. The explanation provided aimed to creatively interpret and apply geometric principles under the constraints of the given information. However, the direct calculation for the diameter of the circumcircle of $\triangle A B D$ requires additional, precise information about the positions and relationships of the points $A$, $B$, $C$, $D$, and $E$ within the quadrilateral and its circumcircle. \\
\caption{An example problem and solution from the OlympiadBench \citep{he2024olympiadbench} in ProcessBench \citep{zheng2024processbench}.}
\label{tab:qwq-zeroshot-example-in}
\end{tcolorbox}
\end{figure}

\pagebreak 

\begin{figure}[]
\begin{tcolorbox}[colback=blue!1!white, colframe=orange!75!black, title=Example from QwQ-32B-Preview]
\footnotesize
Let's verify step by step: \\
\textbf{Step 0:} The initial step seems to correctly identify the given information in the problem. It mentions the quadrilateral $A B C D$ with angles at $B$ and $D$ summing to $270^\circ$, and the circumcircle of $\triangle A B D$ intersecting $\overline{C D}$ at $E$, distinct from $D$. The lengths $B C = 4$, $C E = 5$, and $D E = 7$ are also noted. This step is accurate in summarizing the problem statement. \\

\textbf{Step 1:} This step attempts to apply properties of cyclic quadrilaterals but seems to get confused about the sum of angles in quadrilateral $A B C D$. It is mentioned that in a cyclic quadrilateral, opposite angles sum to $180^\circ$, but the given sum of angles $\angle B$ and $\angle D$ is $270^\circ$, which indeed seems inconsistent. However, the step then suggests assuming that one of the angles is $90^\circ$ for the sake of explanation, which is an adjustment to proceed with the solution. This part is a bit problematic because it alters the given information, but perhaps there was a misinterpretation in the problem statement. \\

\textbf{Step 2:} Here, the Power of a Point Theorem is applied to points related to the circle. However, there is confusion in calculating the length of $\overline{C D}$. It states that $C D = B C + C E = 4 + 5 = 9$, but since $E$ is a point on $\overline{C D}$ distinct from $D$, and $D E$ is given as $7$, this suggests that $C D = C E + D E = 5 + 7 = 12$, not $9$. This is an error in calculation. Furthermore, the application of the Power of a Point Theorem seems misapplied because $E$ is on the circle, not outside it, so the theorem should be applied differently, possibly considering intersecting chords. \\

\textbf{Step 3:} Recognizing the discrepancy in the previous step, the solution acknowledges the need for reevaluation, which is a good practice. However, without correcting the earlier mistake in calculating $\overline{C D}$, the subsequent steps may be built on incorrect foundations. \\

\textbf{Step 4:} This step aims to find the diameter of the circumcircle of $\triangle A B D$, which is the main goal of the problem. It notes that the diameter is related to the circumradius, but without specific details about the triangle's sides or angles, it's challenging to compute directly. The solution hints at the need for more information, which is accurate. \\

\textbf{Step 5:} The insight here is that the relationship between $E$ and the other points might provide the necessary information through chord properties or trigonometric relationships. However, due to the earlier discrepancies, this step does not lead to a concrete calculation. \\

\textbf{Step 6:} The conclusion reflects on the complexities and discrepancies in the problem and the solution approach. It acknowledges that the direct calculation for the diameter is not straightforward with the given information, which is a fair assessment. \\

Overall, while the solution attempt shows an understanding of relevant geometric principles, there are significant errors in calculation and application of theorems, particularly in Step 2. These errors propagate through the subsequent steps, leading to an inconclusive solution. Therefore, the solution is not correct. \\

\text{\textbackslash boxed}\{\text{no}\}

\caption{Example verification chain generated by QwQ-32B-Preview in response to the problem-solution pair in \Cref{tab:qwq-zeroshot-example-in}.}
\label{tab:qwq-zeroshot-example-out}

\end{tcolorbox}

\end{figure}

\pagebreak

\section{More details on LLM-as-judge verification}
\label{app:laj-details}
The instruction used for LLM-as-judge verification in \Cref{sec:rq1} is shown in \Cref{tab:zeroshot-prompt}, which we refer to as Single-label. \Cref{tab:different-prompts} shows performance of QwQ-32B-Preview and Qwen-Deepseek-R1-14B with two more instructions. \textbf{Bad-step-index} refers to instructing the model to find the index of the first incorrect step and then if a step is found generate a single yes/no label regarding the solution correctness. \textbf{Decision-for-each-step} instructs the model to critique and produce a boxed label for each step. This instruction is shown in \Cref{fig:prompt-data-generation}. We observe high ha variance in verification performance and invalid judgment ratio depending on the prompt.
\begin{figure}[]
\footnotesize
\begin{tcolorbox}[colback=blue!1!white, colframe=orange!75!black, title=LLM-as-a-judge prompt]
The following is a math problem and a solution (split into steps, enclosed with tags and indexed from 0):\\

\textbf{[Math Problem]} \\
\{problem\} \\

\textbf{[Solution]} \\
\{prefix\} \\

Your task is to review and critique the solution step by step. Your output must be in the following format:\\

Step 0: \textless critique\textgreater \\
Step 1: \textless critique\textgreater \\
\ldots \\
Step n: \textless critique\textgreater \\

After your analysis, return \textbf{yes/no} depending on whether the solution is correct.\\
Please put your final answer (i.e., yes/no) in \textbackslash boxed\{\}
\end{tcolorbox}
\caption{Prompt used for LLM-as-a-judge verification in \Cref{sec:rq1}}
\label{tab:zeroshot-prompt}
\end{figure}

We also an example from ProcessBench in \Cref{tab:qwq-zeroshot-example-in} and the generated verification chain from QwQ-32B-Preview in \Cref{tab:qwq-zeroshot-example-out}.

\section{Examples from \model}
\label{app:thinkprm-examples}

\Cref{fig:thinkprm-example-verification} shows an example of a verification CoT from \model-14B on a problem and solution from the OmniMath subset in ProcessBench, shown in \Cref{fig:thinkprm-problem-sol}.

\begin{figure}
\footnotesize
\caption{Problem and solution from the OmniMath subset of ProcessBench. The verification CoT output by \model is shown in \Cref{fig:thinkprm-example-verification}}
\begin{tcolorbox}[colback=blue!0!white, colframe=orange!75!black, title=Example verification chain-of-thought]
\label{fig:thinkprm-problem-sol}

\textbf{Problem:} A solid rectangular prism has dimensions 4 by 2 by 2. A 1 by 1 by 1 cube is cut out of the corner creating the new solid shown. What is the surface area of the new solid?

\textbf{Solution:}

\textbf{Step 1:} To determine the surface area of the new solid after cutting a 1 by 1 by 1 cube from one corner of a 4 by 2 by 2 rectangular prism, we will follow these steps: First, calculate the surface area of the original rectangular prism. The original rectangular prism has dimensions 4 by 2 by 2. The surface area \(A_{\text{original}}\) of a rectangular prism is given by:
\[ A_{\text{original}} = 2(lw + lh + wh) \]
where \(l = 4\), \(w = 2\), and \(h = 2\). Substituting these values in, we get:
\[ A_{\text{original}} = 2(4 \cdot 2 + 4 \cdot 2 + 2 \cdot 2) = 2(8 + 8 + 4) = 2 \cdot 20 = 40 \]

\textbf{Step 2:} Second, determine the effect of cutting out the 1 by 1 by 1 cube. When we cut out a 1 by 1 by 1 cube from one corner, we remove three faces of the cube (one face for each dimension) but add three new faces of the same size. Each face of the cube is 1 by 1, so the area of each face is 1 square unit.

\textbf{Step 3:} Third, calculate the change in surface area. Removing the three faces of the cube reduces the surface area by \(3 \times 1 = 3\) square units. Adding the three new faces increases the surface area by \(3 \times 1 = 3\) square units. Therefore, the net change in surface area is:
\[ 3 - 3 = 0 \]

\textbf{Step 4:} Finally, calculate the total surface area of the new solid. Since the net change in surface area is zero, the surface area of the new solid remains the same as the original surface area:
\[ A_{\text{new}} = A_{\text{original}} = 40 \]
Thus, the surface area of the new solid is \(\boxed{40}\).

\end{tcolorbox}
\end{figure}

\begin{figure}
\footnotesize
\begin{tcolorbox}[colback=blue!0!white, colframe=orange!75!black, title=Example verification chain-of-thought from \model-14B]
\textbf{Verification:}
\textcolor{orange}{\texttt{<think>}} \\
Alright, I've got this math problem about finding the surface area of a solid after cutting out a small cube from a larger rectangular prism. The original prism has dimensions 4 by 2 by 2, and a 1 by 1 by 1 cube is removed from one corner. The provided solution has four steps, and I need to review each one to see if it's correct.

Let's start with Step 1.

\textbf{Step 1:} The solution calculates the surface area of the original rectangular prism. The formula used is \( A_{\text{original}} = 2(lw + lh + wh) \), which is correct for a rectangular prism. Plugging in the dimensions \( l = 4 \), \( w = 2 \), and \( h = 2 \), the calculation goes like this:

\[ A_{\text{original}} = 2(4 \cdot 2 + 4 \cdot 2 + 2 \cdot 2) = 2(8 + 8 + 4) = 2 \cdot 20 = 40 \]

This seems right. So, Step 1 is \textbf{\textbackslash\textbackslash{boxed}\{correct\}}.

\textbf{Step 2:} Here, the solution talks about the effect of cutting out the 1 by 1 by 1 cube from one corner. It says that removing the cube takes away three faces but adds three new faces. Each face is 1 by 1, so each has an area of 1 square unit.

Wait a minute, is this accurate? When you cut out a cube from a corner, you are indeed removing material, but how does that affect the surface area?

Let me visualize this. Imagine a rectangular prism and cutting out a small cube from one corner. The cube has three faces exposed to the outside of the prism. When you remove the cube, those three faces are no longer part of the surface area. However, the removal also creates three new internal faces where the cube was taken out, which are now part of the surface area.

So, the solution says that removing three faces reduces the surface area by 3 square units and adding three faces increases it by 3 square units, leading to no net change. That seems logical at first glance.

But is this actually correct? Let me think deeper.

When you remove the cube, you are taking away three 1x1 faces from the original prism, which indeed reduces the surface area by 3 square units. However, the new faces created are also 1x1, and there are three of them, so that adds 3 square units back. So, the total surface area remains the same.

But I'm not entirely sure. Maybe I should consider if the new faces are actually exposed or if they are somehow internal.

Wait, no. When you cut out the cube from the corner, the three new faces are now part of the exterior of the new solid, so they should be included in the surface area.

Therefore, the net change is indeed zero, and the surface area remains 40.

Hmm, but I recall that sometimes when you make a cut, the surface area can change in unexpected ways. Maybe I should double-check this.

Let me consider a simpler example. Suppose I have a cube and I cut out a smaller cube from one corner. How does the surface area change?

Original cube: surface area is \(6s^2\), where \(s\) is the side length.

After cutting out a smaller cube from one corner, I remove three faces of the smaller cube and expose three new faces.

So, the change in surface area is \(-3s^2 + 3s^2 = 0\), similar to what the solution says. \\ 

Alternatively, perhaps I can think about the new solid's dimensions.

But I think the initial approach is correct.

Therefore, Step 2 is \textbf{\textbackslash\textbackslash{boxed}\{correct\}}. \\ 

\textbf{Step 3:} This step calculates the net change in surface area as zero, which follows from Step 2. \\ 

Since the net change is zero, the surface area remains the same as the original.

So, Step 3 is \textbf{\textbackslash\textbackslash{boxed}\{correct\}}.

\textbf{Step 4:} Concludes that the surface area of the new solid is 40, the same as the original.

Therefore, Step 4 is \textbf{\textbackslash\textbackslash{boxed}\{correct\}}.

Wait, but I recall that sometimes when you make a cut, especially in certain ways, the surface area can change. But in this case, it seems to hold up.

\end{tcolorbox}
\end{figure}

\begin{figure}
\footnotesize
\caption{Example output by \model-14B, where it verifies and labels every step in the provided prefix via a long verification CoT. The problem and solution are in \Cref{fig:thinkprm-problem-sol}}
\begin{tcolorbox}[colback=blue!0!white, colframe=orange!75!black]
\label{fig:thinkprm-example-verification}
Alternatively, maybe I should consider if the cube is cut from a corner where only two faces are on the exterior.

Wait, no. In a rectangular prism, each corner is where three faces meet, so cutting out a cube from a corner would remove three faces and expose three new ones.

But perhaps I should consider if the cube is cut from a corner where one of the original faces is larger than 1x1.

Wait, the original prism is 4x2x2, and the cube being removed is 1x1x1, so the faces being removed are indeed 1x1.

So, the calculation seems accurate.

Alternatively, maybe I can think about it in terms of the total surface area.

Alternatively, perhaps I can think about the surface area of the new solid directly.

Let me try that.

The new solid has the original surface area minus the three 1x1 faces that were removed, plus the three new 1x1 faces that are now exposed.

So, 40 - 3 + 3 = 40.

Alternatively, perhaps I can think about the surface area as the sum of all exterior faces.

But I think the initial approach is correct.

Therefore, the solution is correct.

But the instruction says to review and critique each step, and determine if the whole solution is correct.

So, in this case, all steps seem correct, leading to the conclusion that the surface area remains 40.
\textcolor{orange}{\texttt{</think>}}

\end{tcolorbox}
\end{figure}

\begin{figure}[]
\footnotesize
\begin{tcolorbox}[colback=blue!0!white, colframe=orange!75!black, title=Verification of long CoT solution]

\textbf{Problem:}

Convert the point \( (0, 3) \) in rectangular coordinates to polar coordinates. Enter your answer in the form \( (r,\theta) \), where \( r > 0 \) and \( 0 \le \theta < 2\pi \).

\textbf{\\Solution:}

Okay, so I need to convert the rectangular coordinates \( (0, 3) \) to polar coordinates. Let me recall how to do this. 

From what I remember, polar coordinates are given as \( (r, \theta) \), where \( r \) is the distance from the origin to the point, and \( \theta \) is the angle made with the positive \( x \)-axis. The formulas to convert between rectangular \( (x, y) \) and polar \( (r, \theta) \) coordinates are:

\[
x = r \cos \theta, \quad y = r \sin \theta
\]

To find \( r \), we use:

\[
r = \sqrt{x^2 + y^2}
\]

To find \( \theta \), we use:

\[
\theta = \arctan\left(\frac{y}{x}\right)
\]

But I need to be careful with the quadrant since \( \arctan \) may be undefined or ambiguous when \( x = 0 \).

Given \( x = 0 \) and \( y = 3 \), compute:

\[
r = \sqrt{0^2 + 3^2} = \sqrt{9} = 3
\]

So \( r = 3 \). For \( \theta \), since \( x = 0 \), the expression \( \arctan(y/x) \) is undefined due to division by zero.

However, because the point lies on the positive \( y \)-axis, \( \theta = \frac{\pi}{2} \). This is the standard angle for points directly above the origin.

Therefore, the polar coordinates are \textbf{\textbackslash boxed\{(3, \textbackslash frac\{\textbackslash pi\}\{2\})\}}.

Wait, let me check if there's another way to get this. Since the point is \( (0, 3) \), which is on the positive \( y \)-axis, the angle is indeed \( \frac{\pi}{2} \).

Could \( \theta \) be \( \frac{3\pi}{2} \)? No, \( \frac{3\pi}{2} \) is the negative \( y \)-axis, which is the opposite direction. Since the point is on the positive \( y \)-axis, \( \theta = \frac{\pi}{2} \).

Let me recall again that when converting from rectangular to polar, \( \theta \) is calculated as \( \arctan(y/x) \), but when \( x = 0 \), \( \theta \) is either \( \frac{\pi}{2} \) or \( \frac{3\pi}{2} \) depending on the sign of \( y \). Here \( y > 0 \), so the correct angle is \( \frac{\pi}{2} \).

... 

\textbf{\\ \model-1.5B:}
\textcolor{orange}{\\ \texttt{<think>}} \\
\textbf{Step 1}: The student correctly recalls the formulas for converting from rectangular to polar coordinates:
\[
x = r \cos \theta, \quad y = r \sin \theta
\]
and
\[
r = \sqrt{x^2 + y^2}, \quad \theta = \arctan\left(\frac{y}{x}\right)
\]
This step is \textbf{\textbackslash boxed\{correct\}}.

\textbf{Step 2}: The student computes:
\[
r = \sqrt{0^2 + 3^2} = 3
\]
This is \textbf{\textbackslash boxed\{correct\}}.

\textbf{Step 3}: The student identifies that \( \theta = \arctan(3/0) \) is undefined but correctly infers that the point lies on the positive \( y \)-axis, hence \( \theta = \frac{\pi}{2} \). This reasoning is also \textbf{\textbackslash boxed\{correct\}}.

...

\textcolor{orange}{\texttt{</think>}} \\

\end{tcolorbox}
\caption{Example of verification of long CoT solution from Qwen3-1.7B-thinking using \model-1.5B. \model extracts and verifies individual steps embedded in the long CoT.}
\label{fig:qwen3-cot}
\end{figure}

\end{document}